\documentclass[5p, times, authoryear]{elsarticle}
\usepackage{graphicx}
\usepackage{hyperref}
\usepackage{adjustbox}
\usepackage{amsmath}
\usepackage{float}
\usepackage{siunitx}
\usepackage{tabularray}
\usepackage{caption}
\usepackage{lineno}
\usepackage{longtable}

\newcommand{\specialcell}[2][c]{
  \begin{tabular}[#1]{@{}l@{}}#2\end{tabular}}

\UseTblrLibrary{booktabs}
\UseTblrLibrary{diagbox}

\newcommand{\SIadj}[2]{\SI[number-unit-product={\text{-}}]{#1}{#2}}
\newcommand{\SIadjrange}[3]{\SIrange[number-unit-product={\text{-}}]{#1}{#2}{#3}}

\begin{document}

\begin{frontmatter}
  \title{Learning with less: label-efficient land cover classification at very high spatial resolution using self-supervised deep learning\tnoteref{tn1}}
  \author[a]{Dakota Hester\corref{cor1}}
  \ead{dh2306@msstate.edu}
  \author[a]{Vitor S. Martins\corref{cor1}}
  \ead{vmartins@abe.msstate.edu}
  \author[a,b]{Lucas B. Ferreira}
  \author[a]{Thainara M. A. Lima}

  \affiliation[a]{organization={Department of Agricultural and Biological Engineering, Mississippi State University},
    addressline={130 Creelman Street},
    city={Mississippi State},
    postcode={39762},
    state={MS},
    country={USA}
  }
  \affiliation[b]{organization={North Mississippi Research and Extension Center, Mississippi State University},
    addressline={5421 MS-145},
    city={Verona},
    state={MS},
    postcode={38879},
    country={USA}
  }

  \cortext[cor1]{Corresponding author}
  \date{\today}

  \begin{abstract}
Deep learning semantic segmentation methods have shown promising performance for very high \SIadj{1}{\meter} resolution land cover classification, but the challenge of collecting large volumes of representative training data creates a significant barrier to widespread adoption of such models for meter-scale land cover mapping over large areas.
In this study, we present a novel label-efficient approach for statewide \SIadj{1}{\meter} land cover classification using only 1,000 annotated reference image patches with self-supervised deep learning.
We use the “Bootstrap Your Own Latent” pre-training strategy with a large amount of unlabeled color-infrared aerial images (377,921 patches of 256\(\times\)256 pixels at \SIadj{1}{\meter} resolution) to pre-train a ResNet-101 convolutional encoder.
The learned encoder weights were subsequently transferred into multiple deep semantic segmentation architectures (FCN, U-Net, Attention U-Net, DeepLabV3+, UPerNet, PAN), which were then fine-tuned using very small training dataset sizes with cross-validation (250, 500, 750 patches).
Among the fine-tuned models, we obtained 87.14\% overall accuracy and 75.58\% macro F1 score using an ensemble of the best-performing U-Net models for comprehensive \SIadj{1}{\meter}, 8-class land cover mapping, covering more than 123 billion pixels over the state of Mississippi, USA.
Detailed qualitative and quantitative analysis revealed accurate mapping of open water and forested areas, while highlighting challenges in accurate delineation between cropland, herbaceous, and barren land cover types.
These results show that self-supervised learning is an effective strategy for reducing the need for large volumes of manually annotated data, directly addressing a major limitation to high spatial resolution land cover mapping at scale.
\end{abstract}


  \tnotetext[tn1]{
    This manuscript has been accepted for publication in \textit{Science of Remote Sensing} and is available online at \url{https://doi.org/10.1016/j.srs.2026.100397}.
    This accepted manuscript is made available for non-commercial, non-derivative use in accordance with the publisher's policies and the CC BY-NC-ND 4.0 license.
  }

\end{frontmatter}

\section{Introduction}
Land cover data products provide valuable information about the characteristics of the Earth's surface for a wide variety of applications, including agriculture, urban planning, and environmental monitoring~\citep{scarpace1980land,townshendLandCover1992,wulderLandCover2018,banGlobalLandCover2015}.
Research into automated methods for land cover classification from remotely sensed data has been ongoing for several decades, with a majority of methods relying on supervised learning techniques to classify land cover using multispectral aerial or satellite imagery~\citep{brownLessonsLearnedImplementing2020,brownDynamicWorldRealtime2022,friedlMODISCollection52010,zhuContinuousChangeDetection2014}.
Many early efforts into automated land cover classification relied on parametric models to classify individual pixels in coarse resolution satellite imagery based on their spectral signatures or temporal trends~\citep{defriesNDVIderivedLandCover1994,jensenSpectralTexturalFeatures1979,townshendCharacterizationClassificationSouth1987,marceauEvaluationGreylevelCooccurrence1990}.
Pixel-wise classification approaches matured in the late 1990s and 2000s as computational resources became more widely available and more capable non-parametric machine learning models were adopted for such tasks~\citep{lovelandDevelopmentGlobalLand2000,brodleyDecisionTreeClassification1997,palRandomForestClassifier2005,huangAssessmentSupportVector2002,friedlGlobalLandCover2002}.
However, despite the increased predictive capacity of these non-parametric models, classification of very high spatial resolution (VHSR, <\SIadj{5}{\meter}) remained challenging, primarily due to the inherent complexity of land cover patterns at these scales and the increased spectral variability within classes~\citep{strahlerNatureModelsRemote1986,hansenReviewLargeArea2012}.
\citet{woodcockFactorScaleRemote1987} demonstrated that local spectral variability in land cover classes substantially increases as the spatial resolution of imagery increases beyond \SIadj{5}{\meter}, increasing the inherent intra-class variance and making pixel-wise classification of VHSR imagery challenging~\citep{cleveClassificationWildlandUrban2008}.
This is problematic, as VHSR land cover products offer detailed information about the Earth's surface with high spatial accuracy, which is valuable for applications that require fine-grained information about land cover types and their spatial distribution~\citep{liImpactsSpatialResolutions2022,fisherImpactSatelliteImagery2018,heVeryFineSpatial2023}.

Object-based image analysis (OBIA) paradigms emerged in the early 2000s in an effort to address the shortcomings of pixel-wise methods by aggregating neighboring pixels with similar characteristics into objects using a segmentation algorithm; instead of classifying individual pixels, the classification algorithm operates on the objects~\citep{blaschkeObjectBasedImage2010,blaschkeObjectOrientedImageProcessing2000}.
By incorporating spatial context into the classification process, OBIA land cover classification algorithms can outperform their pixel-based counterparts at high spatial resolutions~\citep{plattEvaluationObjectOrientedParadigm2008,balhaComparativeAnalysisDifferent2021,tehranyComparativeAssessmentObject2014,kimMultiscaleGEOBIAVery2011a}.
However, the performance of OBIA classifiers heavily relies on the segmentation algorithm's ability to meaningfully delineate objects in the source imagery in an accurate and reliable fashion.
Selecting and calibrating a segmentation algorithm to robustly segment objects is an ill-defined process that requires substantial human supervision before a supervised classifier can be trained~\citep{maReviewSupervisedObjectbased2017}.
A poor choice of segmentation algorithm or its parameters can substantially degrade performance, as objects extracted by the segmentation algorithm may group multiple objects with different land cover classes together or may fragment a single object into multiple segments~\citep{hossainSegmentationObjectBasedImage2019}.
While multi-scale segmentation algorithms attempt to address these shortcomings, they introduce additional complexity into the classification process that makes developing and scaling these models for operational land cover mapping difficult~\citep{dragutAutomatedParameterisationMultiscale2014,mingScaleParameterSelection2015,haoSegmentationScaleEffect2021}.

Subsequently, the introduction of deep learning-driven convolutional neural networks (CNNs) for general-purpose image processing tasks in the broader computer vision research community sparked substantial interest from researchers in remote sensing looking to consolidate the multi-stage classification process of OBIA techniques into an end-to-end modeling framework capable of incorporating both spectral and spatial features into predictions.
CNNs utilize stacked convolutional kernels with learnable parameters to extract spatially-organized feature maps corresponding to the presence of spectral and spatial features in the input~\citep{krizhevskyImageNetClassificationDeep2012}.
Originally designed for image classification,~\citet{longFullyConvolutionalNetworks2015} demonstrated that CNNs could generate dense pixel-wise classification maps by upsampling these feature maps to match the resolution of the input, then applying a simple linear classifier to each pixel in the upsampled feature maps.
Despite the relative simplicity of this approach compared to existing methods, these fully convolutional networks (FCNs) achieved high performance on difficult general-purpose semantic segmentation tasks.
The design of FCNs became a starting point for research into deep learning models designed explicitly for semantic segmentation, and works centered around integrating these models for classification of remotely sensed data soon followed~\citep{yuanReviewDeepLearning2021}.
These semantic segmentation networks have quickly become the state-of-the-art for VHSR land cover classification~\citep{sherrahFullyConvolutionalNetworks2016,diakogiannisResUNetdeepLearning2020,wangUNetFormerUNetlikeTransformer2022,liDeepLearningUrban2024,zhaoLandUseLand2023}.

Nevertheless, these models require far more labeled training data compared to traditional approaches, serving as a significant barrier to their widespread adoption for operational land cover mapping at VHSR~\citep{zhaoReviewConvolutionalNeural2024,huTransferringDeepConvolutional2015}.
This need for large amounts of ground-truth data for accurate training of deep learning models has led to extensive research into methods for reducing the amount of data necessary to train these model~\citep{wurmSemanticSegmentationSlums2019,tuiaDomainAdaptationClassification2016,martinsDeepLearningHigh2022}.
Most of these methods leverage transfer learning procedures, where the parameters of a model pre-trained using a large dataset are used to initialize a model that is subsequently fine-tuned on a smaller dataset that is more relevant to the target task~\citep{weissSurveyTransferLearning2016,zhuangComprehensiveSurveyTransfer2021}.
For example, a common approach to transfer learning for land cover classification tasks is to pre-train a model on a large general-purpose image classification dataset (e.g., ImageNet~\citep{russakovskyImageNetLargeScale2015}) and then fine-tune the model on a smaller dataset of interest~\citep{piresdelimaConvolutionalNeuralNetwork2020,marmanisSemanticSegmentationAerial2016,nogueiraBetterExploitingConvolutional2017,zouTransferLearningClassification2018}.
Despite being effective, the domain shift between the source and target datasets can hinder the model's performance, and the extent of this domain shift between general-purpose image classification datasets and VHSR remote sensing datasets is not well understood~\citep{maTransferLearningEnvironmental2024}.
As fully-supervised pre-training requires large-scale labeled datasets, which are relatively scarce in the remote sensing domain for VHSR imagery, there is a need for alternative pre-training strategies that reduce the reliance on labeled data~\citep{wangEmpiricalStudyRemote2023,xuSelfsupervisedPretrainingLargescale2024}.

More recently, self-supervised pre-training methods have been proposed as an alternative to traditional fully-supervised pre-training methods, enabling models to learn useful spatial and spectral information from input imagery through the use of pretext tasks that do not require labeled data~\citep{wuUnsupervisedFeatureLearning2018,tianRethinkingFewShotImage2020,chenSimpleFrameworkContrastive2020}.
These techniques are particularly attractive for remote sensing applications, as large amounts of labeled imagery datasets are often unavailable, but source imagery is widely available due to the growing number of Earth observation satellites and aerial imagery platforms~\citep{luVisionFoundationModels2025}.
However, few studies have rigorously investigated the applicability of these methods to operational land cover mapping at VHSR, especially in scenarios where ground-truth data are scarce.
Most research evaluating self-supervised models pre-trained with remote sensing data uses benchmark VHSR land cover datasets as a proxy to evaluate the quality of pre-trained models, with the ISPRS Potsdam~\citep{2DSemanticLabeling} and Vaihingen~\citep{2DSemanticLabel} datasets seeing wide use for this purpose~\citep{sunRingMoRemoteSensing2023,wangMTPAdvancingRemote2024,guoSkySenseMultiModalRemote2024,wangAdvancingPlainVision2023}.
However, these datasets contain relatively large amounts of labeled training data, and thus may not necessarily be representative of real-world scenarios where no available VHSR land cover data exists and costly manual annotation is required to create such data (e.g., by experience, a single 256\(\times\)256 \SIadj{1}{\meter} image patch can take upwards of \SI{30}{\minute} to label well).
Furthermore, while there is growing momentum in the development of large remote sensing foundation models using self-supervised techniques~\citep{luVisionFoundationModels2025,xiaoFoundationModelsRemote2025}, there is still a critical need to systematically assess how these approaches can enhance VHSR land cover mapping performance in practical contexts, a gap this study directly addresses.

Self-supervised approaches inspired by masked language modeling in natural language processing have been shown to be strong approaches for learning useful feature representations from images without the need for labeled data~\citep{heMaskedAutoencodersAre2022,xieSimMIMSimpleFramework2022}.
Masked image modeling (MIM) methods take advantage of the Vision Transformer (ViT) patching mechanism to randomly mask out a portion of the input image and then train the model to reconstruct the missing patches from the visible patches~\citep{linSSMAESpatialSpectral2023,wangFeatureGuidedMasked2025}.
MIM techniques have become the predominant self-supervised pre-training paradigm for creating remote sensing foundation models, which are large pre-trained models that can be adapted for a wide variety of tasks and domains that ideally outperform models trained from scratch or use more traditional transfer learning approaches~\citep{wangAdvancingPlainVision2023,hongSpectralGPTSpectralRemote2024}.
However, while powerful, ViTs lack the inductive biases towards imagery that make CNNs effective at extracting spatial features from images~\citep{adadiSurveyDataefficientAlgorithms2021}.
This is problematic, as ViTs typically only begin to outperform CNNs when pre-trained on very large datasets~\citep{dosovitskiyImageWorth16x162020}, a practical consideration for researchers and practitioners who must balance both computational resources and data availability when developing models.
On the other hand, contrastive pretext tasks train a model to maximize the similarity of embeddings produced by positive pairs (i.e., two randomly augmented views of the same image) while simultaneously minimizing the distance between negative pairs (i.e., two views from separate images).
While contrastive methods are effective and are not restricted to ViTs like their MIM counterparts, they typically rely on large in-memory batch sizes to effectively learn useful representations due to the restrictions of contrastive loss functions.
For example, ablation studies of the popular SimCLR method demonstrate that a batch size of 1024 or greater is necessary to achieve adequate performance~\citep{chenSimpleFrameworkContrastive2020}.
Subsequently, using traditional contrastive pre-training requires large amounts of computational resources, in particular VRAM, for pre-training to be effective~\citep{chenWhyWeNeed2022}.
Momentum contrast (MoCo) is a variant of contrastive pre-training that uses a queue of embeddings to serve as negative pairs: instead of requiring all negative samples to be held in VRAM, a separate momentum encoder network is used to generate these embeddings that are then used in the contrastive loss.
Crucially, this momentum encoder is continuously updated as a slow-moving average of the network being trained. This framework enables more compute-efficient contrastive pre-training~\citep{heMomentumContrastUnsupervised2020,chenImprovedBaselinesMomentum2020}.

Further, self-distillation techniques, such as Bootstrap Your Own Latent (BYOL)~\citep{grillBootstrapYourOwn2020} and DINO~\citep{caronEmergingPropertiesSelfSupervised2021}, are alternatives to MIM and contrastive pre-training that can be used to learn robust and transferable feature representations without the need for negative samples, large batch sizes, or ViT backbones, which are hard requirements for many other self-supervised learning methods~\citep{chenSimpleFrameworkContrastive2020,chenWhyWeNeed2022}.
Such approaches largely revolve around maximizing the similarity between feature representations of different augmented views of the same input using a teacher-student framework~\citep{heFoundationModelBasedMultimodal2024}.
While the student network is trained directly using standard backpropagation and gradient descent algorithms, the teacher's parameters are updated using a momentum-based rule derived from the student's parameters.
Self-distillation methods are attractive for remote sensing applications due to their inherent flexibility and ability to learn strong feature representations using a variety of backbone architectures, including CNNs~\citep{chenExploringSimpleSiamese2020}.
Given the architectural constraints of MIM, contrastive learning and self-distillation techniques offer compelling, architecture-agnostic solutions for pre-training CNNs on domain-specific, moderate-sized datasets~\citep{jainSelfSupervisedLearningInvariant2022}.

In this study, we developed an innovative label-efficient framework that addresses the challenge of operational land cover mapping at \SIadj{1}{\meter} resolution under label-scarce scenarios through self-supervised learning.
Specifically, we evaluated the use of BYOL and MoCoV2 as pretext tasks for self-supervised learning during pre-training, enabling a ResNet-101 convolutional encoder network~\citep{heDeepResidualLearning2015b} to learn to extract deep features from 377,921 unlabeled 256\(\times\)256 patches of \SIadj{1}{\meter} color-infrared (CIR) aerial imagery without the need for full supervision using labeled data.
We then transferred the learned encoder weights into multiple semantic segmentation architectures (FCN, U-Net, Attention U-Net, DeepLabV3+, UPerNet, PAN) that were subsequently fine-tuned using very small training dataset sizes (250, 500, and 750 256\(\times\)256 patches) using cross-validation.
Evaluating the performance of these models showed that initializing a model with out-of-domain ImageNet weights then applying BYOL pre-training on the in-domain imagery led to increases in accuracy across all other pre-training protocols tested.
We then classified 8 classes of land cover (open water, impervious structures, impervious surfaces, barren land, forest/woody vegetation, herbaceous/low vegetation, cultivated crops, and unclassified (i.e., shadow)) over the entire state of Mississippi, USA using an ensemble of the best-performing models, totaling over 123 billion \SIadj{1}{\meter} pixels.
Validation and accuracy assessment were performed using a spatially independent test set of 25,000 randomly sampled points with manually verified labels to ensure robust and unbiased evaluation of our approach and the resulting land cover map.
Finally, we evaluated the generalization capacity of this self-supervised model ensemble by applying it to 2016 imagery with different characteristics and validating the resulting predicted land cover data against a separate ground truth dataset to show that the model can accurately classify imagery outside of the training dataset distribution.
We demonstrated that these approaches can be used to train deep semantic segmentation models that are accurate, generalizable, and capable of yielding reasonable performance even under extreme low-data scenarios, even when only 3 bands (red, green, and near-infrared) are available in the input imagery.
Our ultimate aim is to reduce practical barriers to producing high spatial resolution land cover maps while contributing to ongoing efforts to evaluate and contextualize the implications of self-supervised learning techniques for remote sensing purposes.

\section{Data sources}\label{sec:study_area}

\subsection{NAIP aerial imagery}

\begin{figure*}
    \centering
    \includegraphics[width=\textwidth]{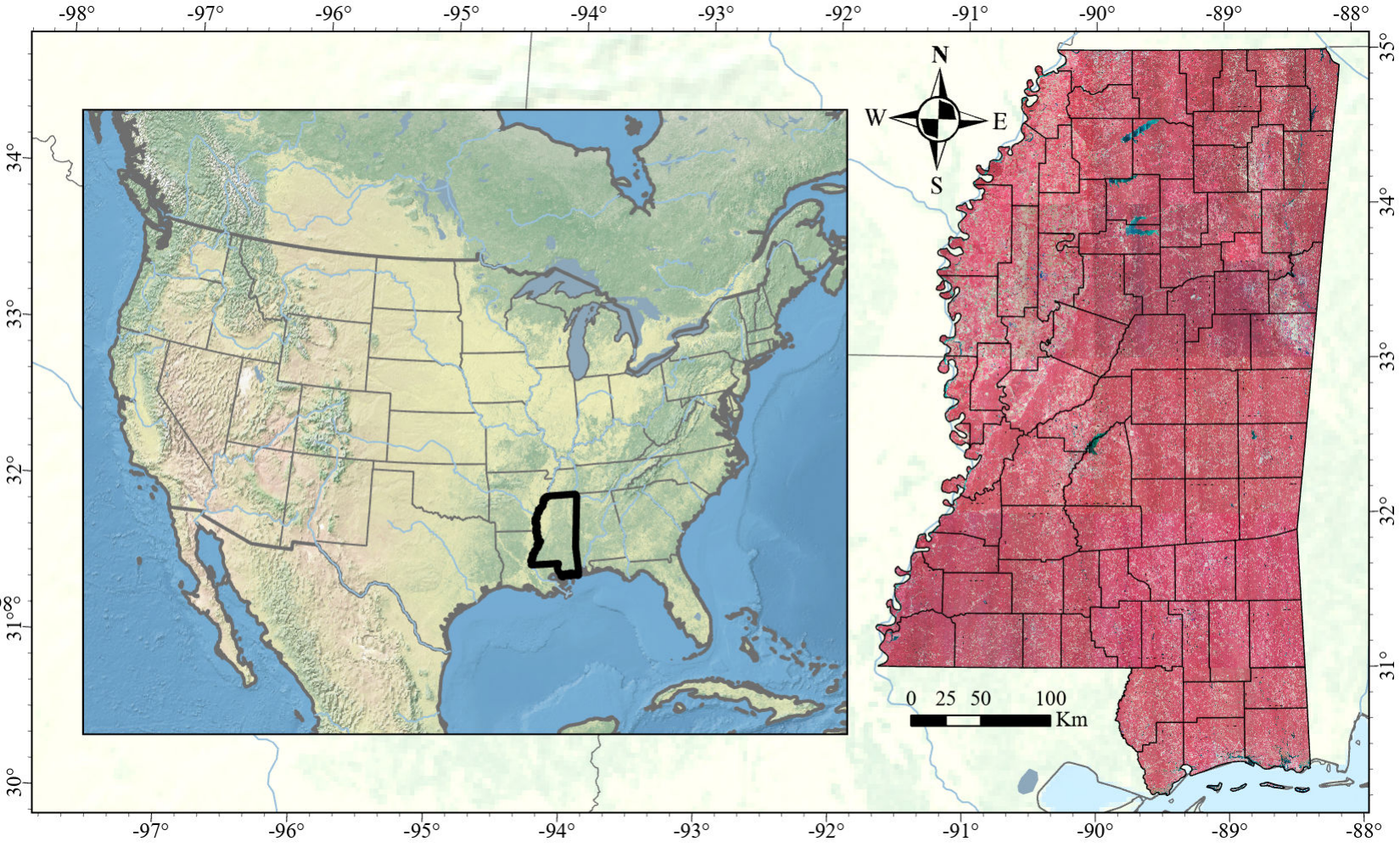}
    \caption{2023 CIR \SIadj{1}{\meter} NAIP imagery over the state of Mississippi, USA that was used in our study for statewide land cover classification..}\label{fig:study_area}

\end{figure*}

The USDA's National Agricultural Imagery Program (NAIP) is a U.S. federal project that provides a publicly available archive of high-resolution multispectral imagery over the conterminous United States, with a spatial resolution of \SIadjrange{0.3}{1}{\meter}, a four-band spectral resolution (near-infrared, red, green, and blue), delivered as 8-bit orthorectified tiles~\citep{earthresourcesobservationandscienceeroscenterNationalAgricultureImagery2017}.
NAIP imagery is collected on a state-by-state basis during the agricultural growing season, and each state is imaged once every 2--3 years. The lack of use restrictions on NAIP imagery has made it a popular choice for high spatial resolution remote sensing applications in the United States, including building footprint extraction~\citep{huangAerialImageryBasedBuilding2022}, forest measurements~\citep{hoglandMappingForestCharacteristics2018,daviesEstimatingJuniperCover2010}, and land cover classification~\citep{maxwellLandCoverClassification2017a,maxwellLargeAreaHighSpatial2019,hayesHighresolutionLandcoverClassification2014,martinsExploringMultiscaleObjectbased2020}.
NAIP imagery is provided as projected rasters according to each raster's appropriate UTM zone, but Mississippi spans across 2 UTM zones (15N/16N); for consistency across the dataset, we reprojected and resampled all imagery to the Mississippi Transverse Mercator coordinate reference system (EPSG:3813) at \SIadj{1}{\meter} resolution using bilinear interpolation. CIR composites for years 2016 and 2023 were generated using the near-infrared, red, and green bands to emphasize vegetation features in the imagery.
We selected three-band images for compatibility with BYOL and MoCoV2, which are designed for three-channel inputs, and for easy comparison with models using ImageNet pre-trained weights, which also expect three-channel inputs.
In total, \SI{1.0}{\tera\byte} of NAIP composite imagery across 2016 and 2023 was downloaded and processed for this study. A visualization of the 2023 NAIP imagery used over the study area is provided in Figure~\ref{fig:study_area}.

\subsection{NLCD land cover}

The stratification approach proposed in this work (Section~\ref{sec:methodology:dataset_curation}) relies on an existing land cover product to select strata from which training samples are drawn. For this purpose, we used the USGS's National Land Cover Database (NLCD) to provide the necessary land cover information for stratification.
The NLCD has historically been among the most widely used land cover datasets in the United States, supporting numerous applications in human and ecosystem health studies, urban planning, natural hazard assessment, hydrology, and land management~\citep{sohlThirtyYearsUS2025}.
The latest version of the NLCD (Annual NLCD) provides a yearly 20-class land cover product primarily derived from Landsat imagery at \SIadj{30}{\meter} resolution from 1985 to 2024~\citep{unitedstatesgeologicalsurveyAnnualNationalLand2025}.
While the spatial resolution of the NLCD land cover product is substantially coarser than the \SIadj{1}{\meter} land cover product we produced in this study, the NLCD provides a comprehensive land cover product at a national scale and has a history of consistent use in the literature~\citep{yangNewGenerationUnited2018,blackardMappingUSForest2008,netzelPatternBasedAssessmentLand2015}, making it a suitable choice to provide land cover information to guide the stratification process.
To align with the collection year of the NAIP imagery used to train the model, we used 2023 data over the state of Mississippi.

\section{Methodology}\label{sec:methodology}

\begin{figure*}
    \centering
    \includegraphics[width=\textwidth]{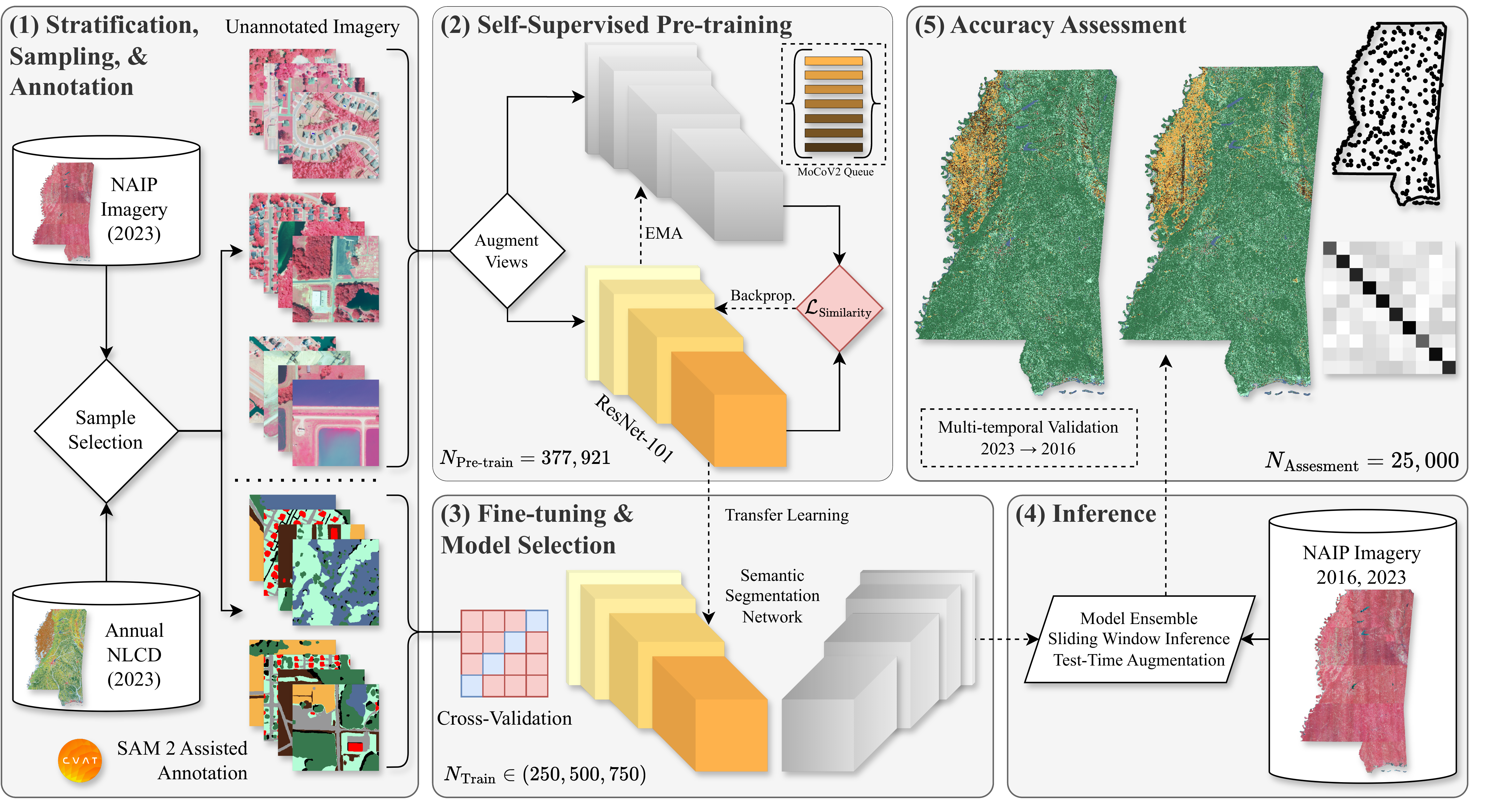}
    \caption{High-level overview of the workflow used to develop, implement, and evaluate a label-efficient strategy for training deep semantic segmentation models for land cover classification over the state of Mississippi with scarce labeled data.}\label{fig:methodology_flowchart}
\end{figure*}

Our proposed methodology followed a multi-stage approach to model development that investigated the implications of self-supervised pre-training with the goal of producing an accurate VHSR land cover product over the state of Mississippi using a limited amount of annotated training data (Figure~\ref{fig:methodology_flowchart}).
In Section~\ref{sec:methodology:dataset_curation}, we describe a sampling strategy designed to capture the diversity of land cover across Mississippi.
Candidate patches were selected and grouped based on their land cover composition, and a subset was sampled for dense visual annotation.
A point-based assessment dataset of 25,000 samples was also created for model evaluation and final product accuracy assessment.
Section~\ref{sec:methodology:self_supervised_pretraining} describes how BYOL and MoCoV2 were used to pre-train a ResNet-101 CNN using 377,921 patches of unannotated pre-training data from the area of interest (excluding the already sampled regions).
Section~\ref{sec:methodology:semantic_segmentation} describes the fine-tuning and evaluation process, where the pre-trained ResNet-101 backbone was integrated into various semantic segmentation architectures and fine-tuned on limited annotated training data using stratified k-fold cross-validation.
We evaluated the performance of multiple segmentation architectures (FCN, U-Net, Attention U-Net, DeepLabV3+, UPerNet, PAN) using varying amounts of training data (250, 500, 750 samples) to understand the implications of pre-training protocol, model architecture, and training dataset size on final model performance.
Several fine-tuned models were compiled into a model ensemble to produce a land cover map at \SIadj{1}{\meter} resolution over the state of Mississippi for the year 2023, with the final data product undergoing a thematic accuracy assessment using the point-based assessment dataset, as described in Section~\ref{sec:methodology:model_ensembling}.
Finally, in Section~\ref{sec:methodology:generalization} we describe how we applied our model to 2016 NAIP imagery to create a multi-temporal land cover product and evaluated the generalization capabilities of the final model ensemble through the use of a separate ground truth validation dataset for 2016 using locations sampled in Section~\ref{sec:methodology:dataset_curation}.

\subsection{Stratified sampling and high-quality land cover training dataset generation}\label{sec:methodology:dataset_curation}

Stratifying the sampling of training data is a common practice in remote sensing to ensure that the training dataset is representative of the land cover distribution of the area of interest, particularly when the area contains rare or underrepresented classes~\citep{stehmanSamplingDesignsAccuracy2009,zhuOptimizingSelectionTraining2016,zhouTrainingDataSelection2020}.
To ensure representative sampling for k-fold cross-validation, we employed a stratified sampling strategy to select 1,000 candidate training samples for annotation.
We overlaid a grid of 256\(\times\)256 \SIadj{1}{\meter} pixels (i.e., 256\(\times\)\SIadj{256}{\meter} patches), forming a population of 1,889,606 candidate patches used to support both the unlabeled dataset for pre-training and the labeled dataset for fine-tuning.
The patch size was chosen following~\citet{diakogiannisResUNetdeepLearning2020} and~\citet{liDKDFNDomainKnowledgeGuided2022}, with 256\(\times\)256 pixels offering sufficient spatial context for supporting high classification accuracy while still remaining computationally efficient~\citep{jaturapitpornchaiNewlyBuiltConstruction2019,reddyEffectHyperparametersDeepLabv32023}.

\begin{figure*}
    \centering
    \includegraphics[width=\textwidth]{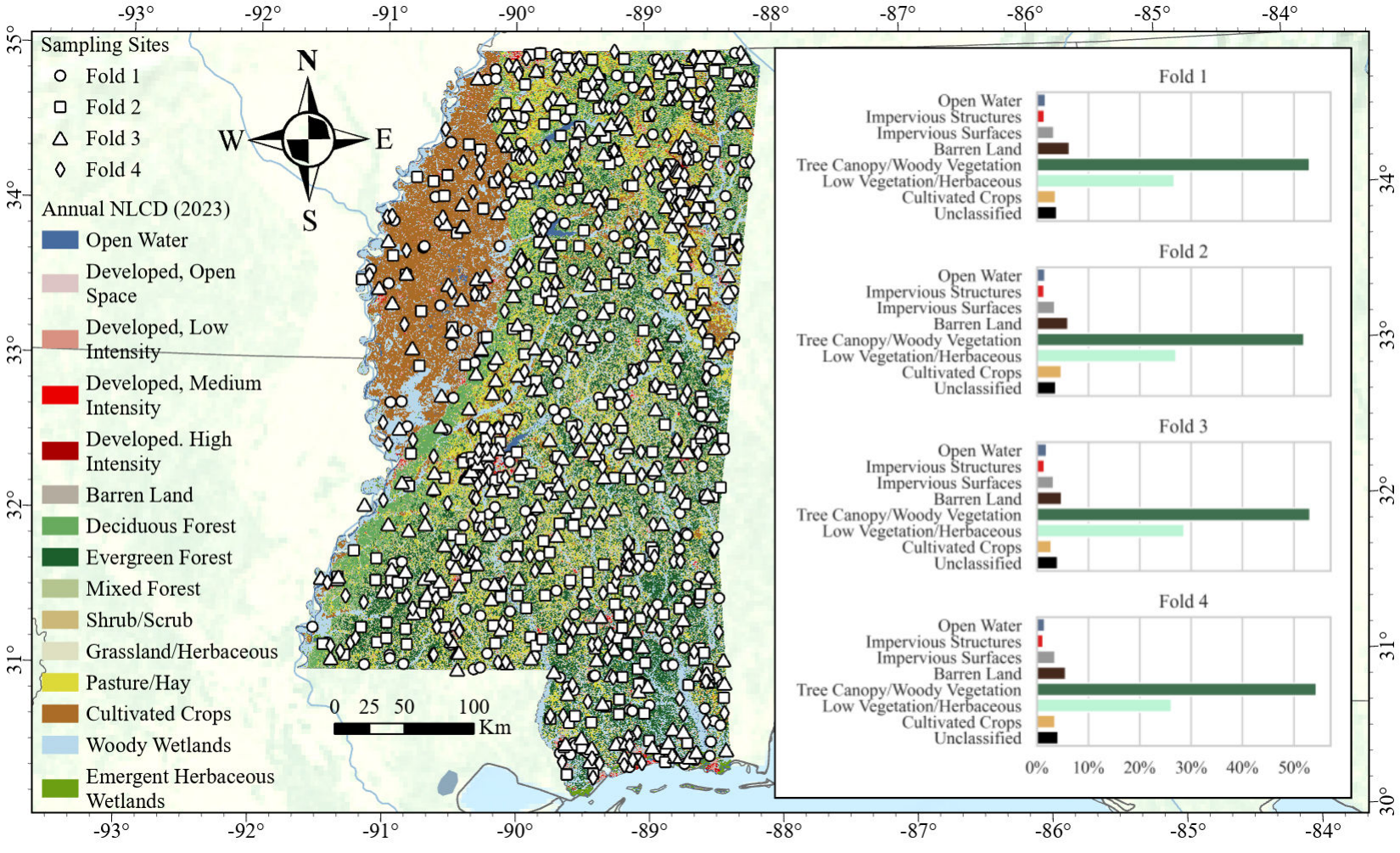}
    \caption{Locations of 1,000 sampled patches overlaid on NLCD land cover data for the state of Mississippi, USA (left), and the distribution of land cover classes within each fold (right).}\label{fig:nlcd_samples}
\end{figure*}

To curate the fine-tuning dataset, zonal histograms of land cover class distribution were extracted from the 2023 NLCD dataset for each candidate patch to characterize its land cover composition as a type of feature vector.
We then applied PCA to reduce the dimensionality of the feature vectors and K-means clustering to form 250 strata based on land cover composition.
Finally, we randomly sampled 4 candidate training samples from each stratum, placing each into a separate fold of the k-fold training dataset, resulting in 4 folds totaling 1,000 samples.
Figure~\ref{fig:nlcd_samples} shows the NLCD land cover across Mississippi, along with the spatial distribution of the selected samples and the class composition within each fold.
A nearest neighbor analysis revealed a statistically significant spatial clustering effect (\(Z=-3.211, p=0.001\)), though this clustering effect is mild (\(R=0.949\)).

\begin{figure*}
    \centering
    \includegraphics[width=\textwidth]{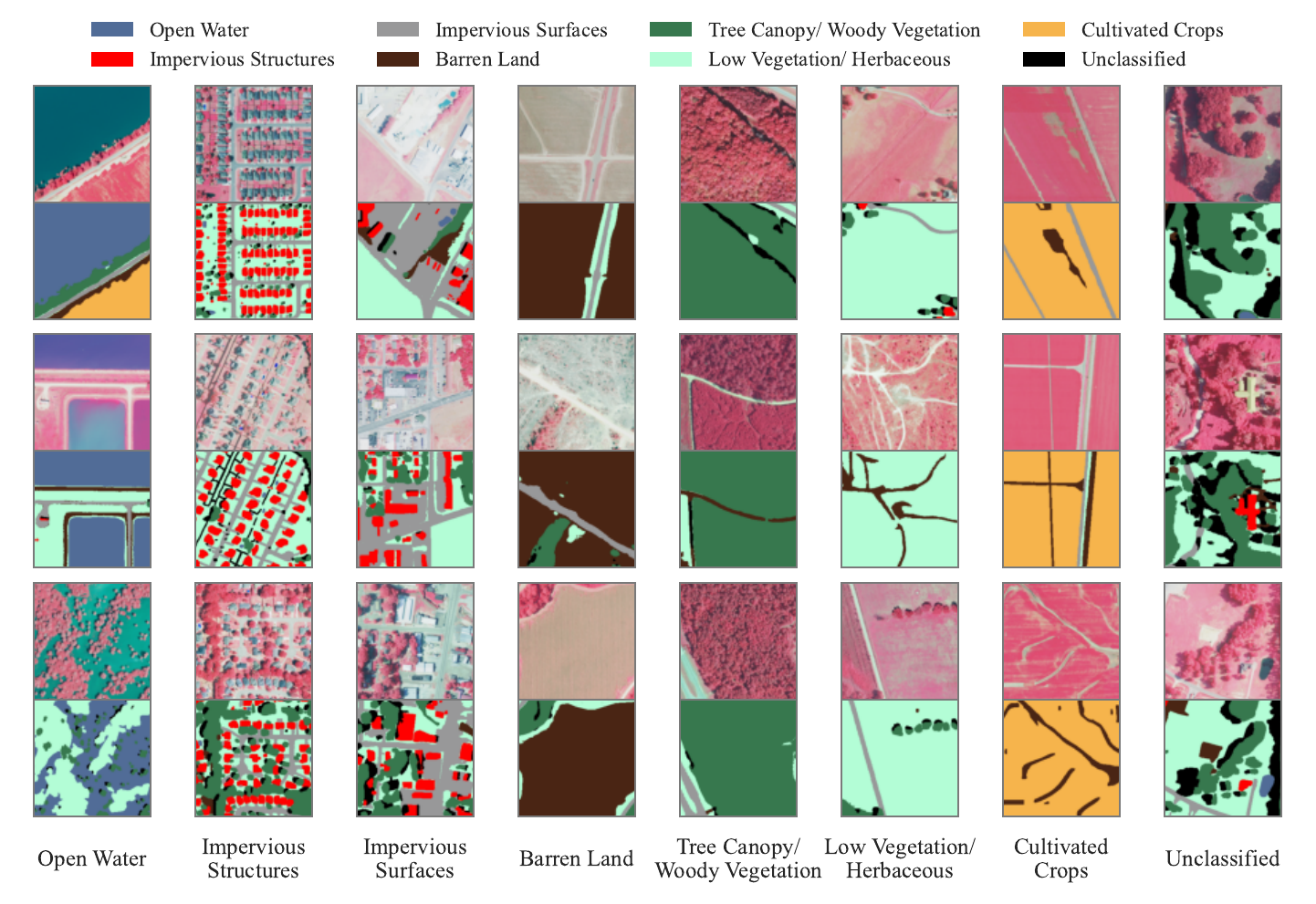}
    \caption{Examples of ground truth samples and corresponding NAIP imagery used for model training.}\label{fig:samples}
\end{figure*}

A team of trained annotators visually annotated the selected 1,000 samples according to the land cover classification scheme in Table~\ref{tab:class_legend}.
Samples were randomly assigned to each annotator to ensure that each fold contained a mix of contributions from different individuals.
After the annotators completed their work, all annotations were reviewed and corrected by a single annotator.
Only one annotator was used for the quality assurance phase of the annotation process to reduce the effects of inter-annotator disagreement and to enforce a consistent, high quality annotation scheme across the entirety of the ground truth dataset.
Annotators employed the Segment Anything Model 2 (SAM 2) hosted on the Computer Vision Annotation Tool (CVAT) annotation platform to assist with the visual annotation process~\citep{cvat2023,raviSAM2Segment2024} by proposing initial object masks that annotators could manually refine as needed.
If SAM 2 was unable to accurately mask an object or feature in the imagery, annotators fell back to manual annotation. Examples of annotated samples are shown in Figure~\ref{fig:samples}.

\begin{table*}
    \centering
    \caption{Land cover classification scheme for the Mississippi dataset.}\label{tab:class_legend}
    \begin{tabular*}{\linewidth}{clp{12cm}}
        \toprule
        \textbf{Index} & \textbf{Class} & \textbf{Description} \\
        \midrule
        1 & Open Water & Rivers, lakes, ponds, and other standing or moving water. Water bodies typically appear dark in CIR imagery due to near-infrared absorption; may appear bluish-green depending on depth and turbidity. \\
        \addlinespace
        2 & Impervious Structures & Elevated built features (e.g., buildings, sheds). Bright and reflective with sharp edges and shadows. \\
        \addlinespace
        3 & Impervious Surfaces & Paved areas such as roads and parking lots. Typically bright but may appear dark depending on material. Linear patterns and lack of shadows distinguish from barren land and impervious structures. \\
        \addlinespace
        4 & Barren Land & Exposed soil, sand, or rock with minimal vegetation. Bright or neutral in CIR imagery, often interleaved with herbaceous areas. \\
        \addlinespace
        5 & Forest/Woody Vegetation & Tree-covered areas with dense canopy. Deep red in CIR imagery due to high near-infrared reflectance. Uneven texture with large shadows. \\
        \addlinespace
        6 & Herbaceous/Low Vegetation & Short vegetation, including grasses and shrubs. Smoother texture than forest, lighter red in CIR imagery. Common in clearings and field edges. \\
        \addlinespace
        7 & Cultivated Crops & Actively farmed fields, often in distinct rows. Bright red or pink in CIR imagery at peak growth. Presence of rows or patterns are used to distinguish from herbaceous areas. \\
        \addlinespace
        8 & Unclassified & Shadows or occlusions obscuring surface features. Deep black regions near trees and buildings. \\
        \bottomrule
    \end{tabular*}
\end{table*}

In addition to the 1,000 samples selected for annotation, a large dataset of unannotated samples was also collected from the state of Mississippi using simple random sampling from the remaining candidate patches to form a large dataset of unannotated samples for pre-training the ResNet-101 encoder.
20\% (377,921) of the total number of candidate samples were selected for pre-training, while 5\% (94,480) were selected to serve as a pre-training validation dataset to monitor for overfitting during the pre-training process.
Once all sample patches were selected, an additional point-based assessment dataset of 25,000 samples was created by randomly sampling points from the remaining candidate samples to serve as a testing dataset for the purposes of model evaluation and final accuracy assessment of the resulting land cover product.
No points within \SI{200}{\meter} of a sampled patch were selected to prevent data leakage caused by spatial autocorrelation, and no points within \SI{1}{\kilo \meter} of a previously sampled point were selected to ensure that the assessment points were spatially independent of each other.
These points were visually annotated by a single annotator using 2023 NAIP imagery. For model evaluation purposes, 256\(\times\)\SIadj{256}{\meter} patches were extracted around each point such that the point was at the center of the patch, and a model's prediction for the central pixel corresponding to the point was compared to the point label to compute accuracy metrics.

\subsection{Self-supervised pre-training}\label{sec:methodology:self_supervised_pretraining}

\begin{figure*}
    \centering
    \includegraphics[width=\linewidth]{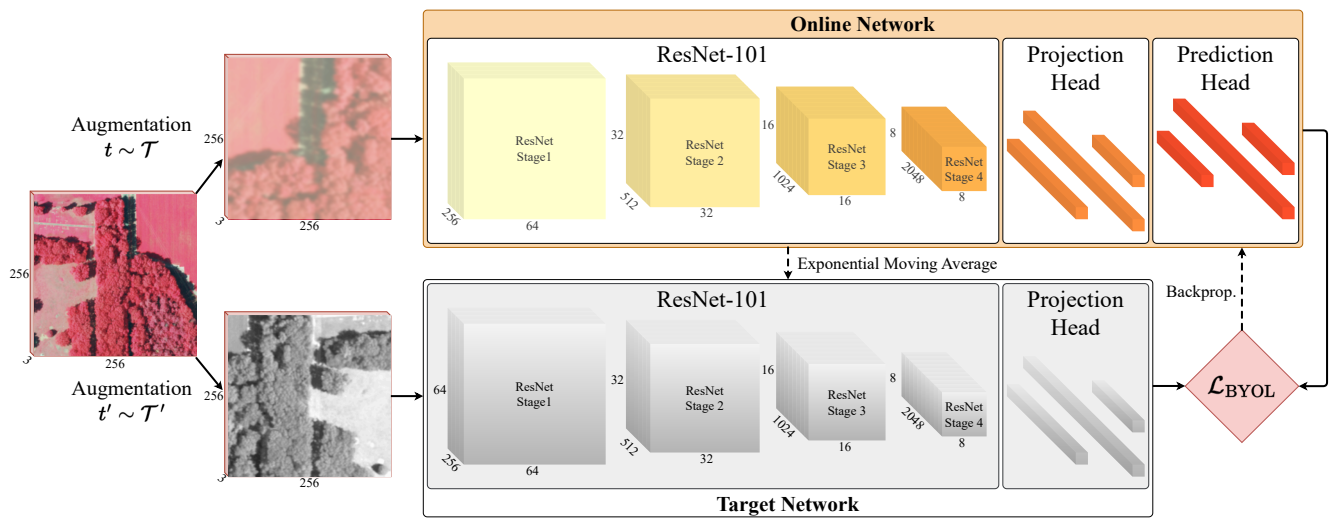}
    \caption{BYOL pre-training step.}\label{fig:byol_workflow}
\end{figure*}

The pre-training phase of the methodology leverages the large dataset (377,921) of unannotated samples such that a ResNet-101 encoder can be trained to extract relevant features from the imagery that are well-suited for downstream land cover classification tasks.
We compared two prominent self-supervised learning approaches: BYOL and MoCoV2.
BYOL uses a self-distillation approach to learn representations from images using two separate but interdependent networks: an online network containing the image encoder, a projection head, and a prediction head, and a target network that is structurally similar to the online network but lacks the prediction head, initialized using a copy of the online network's parameters at the start of training~\citep{grillBootstrapYourOwn2020}.
Given an input image, two separate augmented views of the image are created using a set of transformations. As illustrated in Figure~\ref{fig:byol_workflow}, each view is then passed through the online and target networks to produce feature vectors.
The online network, which processes view \(v_1\), is then trained to predict the feature vector produced by the target network from view \(v_2\).
However, the target network is not trained directly; instead, it is updated using an exponential moving average (EMA) of the online network weights, giving the online network a stable yet constantly updated target to learn from.
Ultimately, the image encoder from the online network is retained after training to be transferred to downstream tasks.

The online encoder's parameters are updated with respect to the loss function in Equation~\ref{eq:byol_loss}, where \(p\) is the output of the online network and \(z\) is the output of the target network.

\begin{equation}\label{eq:byol_loss}
    \mathcal{L}_\text{BYOL} = 2 - 2 \frac{\langle p, z \rangle}{\|p\| \|z\|}
\end{equation}

The target network's parameters \(\theta_\text{target}\) are updated according to the EMA update rule shown in Equation~\ref{eq:ema_update}, where \(\theta_\text{online}\) is the current set of parameters for the online network and \(m\) is the decay factor for the EMA operation, and is increased from 0.996 to 1 over the course of pre-training according to a cosine annealing schedule.

\begin{equation}\label{eq:ema_update}
    \theta_\text{target} \leftarrow m \theta_\text{target} + (1 - m) \theta_\text{online}
\end{equation}

BYOL uses an asymmetrical set of augmentations: the base set of augmentations T consists of random crops (scale \(\in\left[0.08,1.0\right]\), ratio \(\in\left[3/4,4/3\right]\)), horizontal flips (50\% probability), random color jitter (brightness = 0.4, contrast = 0.4, saturation = 0.2, hue = 0.2; 80\% probability), random grayscale conversion (20\% probability), and Gaussian blur (kernel size = 25, \(\sigma\in\left[0.1,2.0\right]\)).
A separate set of augmentations \(T^\prime\) is also used, containing mostly the same set of augmentations as \(T\)) but with an additional random solarization procedure (threshold=0.5, 20\% probability) and reducing the probability of Gaussian blur being applied to 10\% from 100\%.
This asymmetry is enforced during training: given an image, each network will receive a view with augmentations drawn from either \(T\) or \(T^\prime\), but both networks will never receive a view augmented using the same set of augmentations (e.g., both the online and target networks can not receive two randomly augmented images with augmentations in \(T\)).

Unlike BYOL, MoCoV2 is a contrastive learning paradigm that trains a network to maximize the cosine similarity between positive pairs (i.e., pairs of embeddings produced by a network given two random augmentations of the same image) while simultaneously minimizing the cosine similarity between negative pairs (i.e., pairs of embeddings derived from augmentations of different images)~\citep{heMomentumContrastUnsupervised2020,chenImprovedBaselinesMomentum2020}.
MoCoV2 is closely related to SimCLR, a straightforward contrastive pretext task where the set of positive and negative pairs used to calculate loss is derived from a batch of images passed through an encoder network.
While straightforward, contrastive learning works best with large numbers of negative samples, and thus SimCLR (and similar contrastive pre-training techniques) requires large in-memory batches for best performance~\citep{chenSimpleFrameworkContrastive2020,chenWhyWeNeed2022}.
MoCoV2 addresses this by using a queue of negative pairs for the contrastive loss: embeddings are placed in the queue on the fly during training and serve as negative samples until they are eventually removed from the queue.
This enables practical contrastive learning with large amounts of negative pairs, as the queue only needs to hold embeddings, not full images.
To ensure the embeddings in the queue remain consistent despite the ever-evolving nature of the encoder network, a momentum (key) encoder network is used to construct the embeddings that will be enqueued.
Like the teacher network in BYOL, this momentum encoder network is simply a slow moving average of the primary (query) encoder that is being trained.
Unlike BYOL, these networks are symmetrical: each consists of a convolutional backbone (in our case ResNet-101) and a projection head.
Like BYOL, however, the query and key encoders receive different augmentations of the same image.
The loss is calculated according to the InfoNCE loss shown in Equation~\ref{eq:infonce_loss}, where \(q\) is the embedding produced by the query encoder, \(k^+\) is the embedding produced by the key encoder (the positive pair), \(\left\{k^-\right\}\) is the queue of negative samples, and \(\tau\) is the temperature parameter (set to 0.2 per the original MoCoV2 implementation).

\begin{equation}\label{eq:infonce_loss}
    \mathcal{L}_\text{InfoNCE} = -\log \frac{\exp{\left(\langle q, k^+ \rangle / \tau\right)}}{\exp{\left(\langle q, k^+ \rangle / \tau\right)} + \sum_{\{k^-\}} \exp{\left(\langle q, k^- \rangle / \tau\right)}}
\end{equation}

The InfoNCE loss is backpropagated over the query encoder’s parameters, while the key encoder is updated according to Equation~\ref{eq:ema_update}, where \(m=0.999\), \(\theta_{\mathrm{target}}\) corresponds to the key encoder, and \(\theta_{\mathrm{online}}\) corresponds to the query encoder.
This momentum term is fixed throughout training to match the original MoCoV2 implementation.
\(k^+\) is then immediately placed in the queue of negative samples; if the queue size \(K\) exceeds 65,536, the oldest embeddings in the queue are dequeued and discarded.
After training is finished, the backbone of the query encoder is transferred to the downstream task, while the projection head and key encoder are discarded. MoCoV2 draws its set of augmentations from SimCLR, with random crops (scale \(\in\left[0.2,1.0\right]\), ratio \(\in\left[3/4,4/3\right]\)), random horizontal flips (50\% probability), random color jitter (brightness = 0.4, contrast = 0.4, saturation = 0.4, hue = 0.1; 80\% probability), random grayscale conversion (20\% probability), and random Gaussian blur (kernel size = 25, \(\sigma\in\left[0.1,2.0\right]\); 20\% probability) being used to generate augmented views of images.
We applied both BYOL and MoCoV2 pre-training to a ResNet-101 encoder network using the dataset of 377,921 unannotated samples from the state of Mississippi.
We tested two configurations of each pre-training process: one starting with ImageNet pre-trained weights at the beginning of the pre-training process and one starting with randomly initialized weights.
The pre-training process was run for 300 epochs, with the learning rate reduced from \(1\times{10}^{-3}\) to 0 over the course of training using a cosine annealing schedule, with a warmup period of 10 epochs to linearly increase the learning rate from 0 to \(1\times{10}^{-3}\).
The AdamW optimizer was used with a batch size of 4096: to achieve this large batch size while working within the memory constraints of a single Nvidia A100 GPU with 80 GB of VRAM, we employed gradient accumulation using a microbatch size of 256.
After each epoch, the model was evaluated on the pre-training validation dataset using the pretext task's loss function (as during training) to monitor for overfitting.

\begin{figure*}
    \centering
    \includegraphics[width=\textwidth]{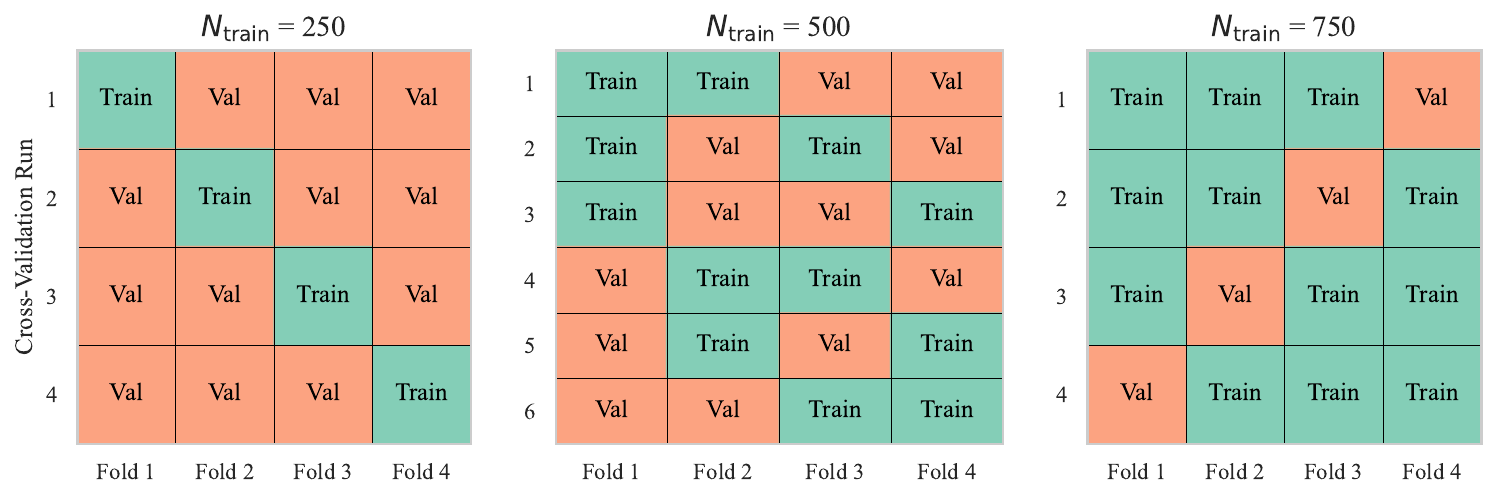}
    \caption{Folds of the training dataset used for k-fold cross-validation under 250, 500, and 750 training samples per fold.}\label{fig:fold_splits}
\end{figure*}

To evaluate the quality of the representations learned during pre-training, we used a linear probing approach to determine how well the deep features extracted by the pre-trained networks could be used to classify land cover classes using a simple linear classifier.
To accomplish this, the feature maps produced by the ResNet-101 encoder were upsampled to the original input resolution using bilinear interpolation, and a linear classifier was trained on top of these feature maps using the annotated training dataset from Section~\ref{sec:methodology:dataset_curation}.
The linear classifier consisted of a single \(1\times1\) convolutional layer with a softmax activation function to produce class probabilities for each pixel.
K-fold cross-validation was employed to minimize the influence of variation in the training dataset while evaluating the performance of this method across different training dataset sizes \(N_\text{train} \in (250,500,750)\); the splitting strategy for each \(N_\text{train}\) is shown in Figure~\ref{fig:fold_splits}.

A separate model was trained for each run of the cross-validation framework, with the model being trained for a maximum of 1,000 epochs using the AdamW optimizer with a learning rate of \(1\times{10}^{-4}\) and a batch size of 32~\citep{loshchilovDecoupledWeightDecay2019}.
Focal loss was used as the loss function to address the class imbalance in the land cover classes as shown in Equation~\ref{eq:focal_loss}, where \(p_t\) is the model's estimated probability for the true class and \(\gamma=2.0\)~\citep{linFocalLossDense2017}.

\begin{equation}\label{eq:focal_loss}
    \mathcal{L}_\text{Focal}(p_t) = -(1 - p_t)^\gamma \log(p_t)
\end{equation}

A warmup period of 10 epochs was used to linearly increase the learning rate from \(1\times{10}^{-5}\) to \(1\times{10}^{-4}\) at epoch 10, after which the learning rate was decayed by a factor of 0.1 when the validation loss failed to decrease for 10 epochs.
After 50 epochs of no improvement in the validation loss, training was halted and the model with the lowest validation loss was selected as the final model for that run. Automatic mixed precision (AMP) training was used to speed up the training stage and reduce memory usage.
Due to the limited size of the training datasets, we employed a strong set of image augmentations on-the-fly to ensure that the models did not overfit to the training data, including random horizontal and vertical flips (50\% probability each), random 90\(^\circ\) rotations (0\(^\circ\), 90\(^\circ\), 180\(^\circ\), 270\(^\circ\)), color jitter (brightness = 0.4, contrast = 0.4, saturation = 0.2, hue = 0.2; 80\% probability), random grayscale conversion (20\% probability), and Gaussian blur (kernel size = 25, \(\sigma\in\left[0.1,2.0\right]\)).
Since each cross-validation run used its validation split to tune the optimizer and trigger early stopping, we did not use validation-split metrics for final performance reporting.
Rather, the performance of the linear probing models was evaluated using the point-based assessment dataset (25,000 samples), with the results being compared between the two MoCoV2 and two BYOL pre-training configurations (i.e., different weight initialization strategies) as well as a baseline model that used the ResNet-101 encoder with ImageNet pre-trained weights without any further pre-training.
Considering the natural imbalance of the land cover classes, we chose the macro F1 score as the primary evaluation metric, as shown in Equation~\ref{eq:macro_f1}, where \(C\) is the number of classes, \(TP_c\) is the number of true positives for class \(c\), \(FP_c\) is the number of false positives for class \(c\), and \(FN_c\) is the number of false negatives for class \(c\).

\begin{equation}\label{eq:macro_f1}
    \text{Macro F1 Score} = \frac{1}{C} \sum_{c=1}^{C} \frac{2\text{TP}_c}{2\text{TP}_c + \text{FP}_c + \text{FN}_c}
\end{equation}

\subsection{Implementation of multiple semantic segmentation architectures with fine-tuning under limited labeled dataset}\label{sec:methodology:semantic_segmentation}

Once the ResNet-101 encoders were pre-trained, they were used as a backbone for a semantic segmentation architecture that was fine-tuned using the annotated training dataset from Section~\ref{sec:methodology:dataset_curation}.
We evaluated 6 different semantic segmentation architectures to determine which framework was best suited for the land cover classification task at hand.
We excluded any architectures that rely on transformer-based encoders (e.g., SegFormer~\citep{xieSegFormerSimpleEfficient2021}, SETR~\citep{zhengRethinkingSemanticSegmentation2021}) or specialized convolutional backbones (e.g., HRNet~\citep{wangDeepHighResolutionRepresentation2020}, BiSeNetV2~\citep{yuBiSeNetV2Bilateral2021}) to maintain a controlled comparison between architectures that all use the same ResNet-101 encoder network.
Table~\ref{tab:semantic_segmentation_models} shows the semantic segmentation architectures that were evaluated in this study, and a brief description of each architecture is provided below.

\begin{table}[ht]
    \centering
    \caption{Semantic segmentation architectures evaluated in this study, with their total parameters using a ResNet-101 backbone.}\label{tab:semantic_segmentation_models}
    \footnotesize
    \begin{tabular*}{\linewidth}{@{\extracolsep{\fill}}lll}
        \toprule
        \textbf{Model} & \textbf{Number of Parameters} & \textbf{Reference} \\
        \midrule
        FCN & 42,528,856 & \citet{longFullyConvolutionalNetworks2015} \\
        U-Net & 92,302,024 & \citet{ronnebergerUNetConvolutionalNetworks2015} \\
        Attention U-Net & 83,610,272 & \citet{oktayAttentionUNetLearning2018} \\
        DeepLabV3+ & 58,199,128 & \citet{chenEncoderDecoderAtrousSeparable2018} \\
        UPerNet & 73,251,400 & \citet{xiaoUnifiedPerceptualParsing2018} \\
        PAN & 140,688,970 & \citet{liPyramidAttentionNetwork2018} \\
        \bottomrule
    \end{tabular*}
\end{table}

\paragraph{FCN}
The Fully Convolutional Network (FCN) was one of the earliest deep learning architectures designed specifically for semantic segmentation tasks~\citep{longFullyConvolutionalNetworks2015}.
As its name suggests, FCN is composed entirely of convolutional layers, eschewing the use of fully connected layers that are common in image classification architectures.
Under this design, instead of outputting a single class label for an entire image, the model outputs a spatial map of class probabilities that correspond to the input image's spatial dimensions.
FCN uses a standard convolutional encoder to extract hierarchical features from the input image, followed by a series of upsampling layers to produce the final segmentation map.
Though simple, FCN was a groundbreaking architecture that demonstrated the potential of deep learning for semantic segmentation tasks, and its encoder-decoder framework has been widely adopted in subsequent architectures.
FCN saw early success in remote sensing applications, particularly for land cover classification tasks at high spatial resolutions~\citep{sherrahFullyConvolutionalNetworks2016, volpiDenseSemanticLabeling2017, chenSymmetricalDenseShortcutDeep2018}.
We implemented the FCN-8s variant of FCN, which uses skip connections to combine feature maps from the encoder at multiple stages to produce more accurate segmentation maps.

\paragraph{U-Net}
U-Net is a popular semantic segmentation architecture that was among the early adopters of the encoder-decoder framework for semantic segmentation tasks.
Under such a framework, the encoder network extracts hierarchical features from the input image, while the decoder network learns a mapping from these low-dimensional features to an output segmentation map that matches the original input resolution~\citep{ronnebergerUNetConvolutionalNetworks2015}.
Notably, U-Net's decoder network is approximately symmetric to the encoder network, meaning that the feature maps are upsampled in stages that mirror the downsampling stages of the encoder.
As such, skip connections can be used to pass feature maps from the encoder to the decoder at corresponding stages: the encoder's feature maps are concatenated with the decoder's feature maps at each stage to provide the decoder with relevant, resolution-specific information that otherwise would be lost during the downsampling process.
U-Net's simple yet powerful architecture has made it a popular choice for a wide variety of remote sensing applications~\citep{robinsonLargeScaleHighresolution2019,yiSemanticSegmentationUrban2019, liDeepUNetDeepFully2018, schieferMappingForestTree2020}
Our U-Net implementation differed from the original U-Net architecture in two key ways: first, we opted to use simple bilinear interpolation for upsampling in the decoder rather than transposed convolutions to reduce the presence of checkerboard artifacts in the output segmentation maps \citep{odenaDeconvolutionCheckerboardArtifacts2016}; second, as we used a ResNet-101 encoder that downsamples the input image by a factor of 4 in the first stage (as opposed to the factor of 2 downsampling in the original U-Net), we modified the decoder to account for this difference by adding an additional upsampling stage to ensure that the output segmentation map matches the input resolution.

\paragraph{Attention U-Net}

As the name suggests, Attention U-Net is a variant of the U-Net architecture that incorporates attention mechanisms that emphasize spatial regions of interest in the feature maps.
This is accomplished through the use of attention gates along the skip connections between the encoder and decoder, which learn to weight the features passed from the encoder to the decoder based on their relevance to the current task.
By focusing on the most relevant spatial regions, \citet{oktayAttentionUNetLearning2018} demonstrated that Attention U-Net achieved improved performance on medical image segmentation tasks compared to the original U-Net architecture, particularly with regard to the localization of object boundaries.
Like U-Net, our Attention U-Net implementation used bilinear interpolation for upsampling in the decoder rather than transposed convolutions, and the decoder included an additional upsampling stage to account for the ResNet-101 encoder's downsampling factor of 4 in the first stage.

\paragraph{DeepLabV3+}

DeepLabV3+ is an evolution of the DeepLab family of semantic segmentation architectures that incorporates atrous (dilated) convolutions and spatial pyramid pooling to capture multi-scale context in the feature maps~\citep{chenEncoderDecoderAtrousSeparable2018}.
At the core of DeepLabV3+ is an Atrous Spatial Pyramid Pooling (ASPP) module that aggregates multi-scale context from the final feature maps produced by the encoder.
This ASPP module uses parallel atrous convolutions to encode feature maps at multiple scales as opposed to using pooling operations: the use of atrous (i.e., dilated) convolutions allows the network to capture features at multiple scales in a learned manner.
A comparatively simple decoder network is utilized that upsamples the output of the ASPP module using bilinear interpolation, concatenates it with low-level feature maps from the encoder, and upsamples the result to produce the final segmentation map.
Like U-Net, DeepLabV3+ has seen wide adoption for remote sensing applications~\citep{duIncorporatingDeepLabv3Objectbased2021, liuComparisonMultisourceSatellite2021,krestenitisOilSpillIdentification2019}.
DeepLabV3+ was originally designed to use a modified ResNet-101 encoder, and as such minimal effort was necessary to replace the original encoder with our pre-trained ResNet-101 encoder.

\paragraph{UPerNet}
Unlike most semantic segmentation architectures that only attempt to classify objects at a single scale, UPerNet (Unified Perceptual Parsing Network) is designed to classify objects as part of a hierarchical structure that operates at multiple levels of granularity.
UPerNet classifies an image at one global level (i.e., scene classification) and at four local levels (object, part, material, and texture classification maps)~\citep{xiaoUnifiedPerceptualParsing2018}.
The core component of UPerNet is Feature Pyramid Network (FPN), a U-Net-like encoder-decoder architecture that serves as the primary feature extractor for UPerNet.
The encoder network's downsampling stages are similar to those of ResNet-101, making substituting the original FPN encoder with a ResNet-101 encoder straightforward.
A pyramid pooling module is placed after the encoder network in the FPN to aggregate global context from the feature maps prior to passing them to the decoder.
Our implementation disregarded the multi-level classification aspect of UPerNet, and only used the object-level classification head to produce a single land cover classification map.
This object-level classification head receives feature maps from all levels of the FPN decoder, classifies each pixel, and upsamples the output to match the input resolution.
Classifying land cover using feature maps from all levels of the decoder allows the classification head to leverage features at multiple scales to inform its predictions, unlike more traditional encoder-decoder architectures that only consider the output from the final, highest-resolution feature map of the decoder.

\paragraph{PAN}

Incorporating multi-scale context into semantic segmentation models is a common theme among many modern architectures, but many of these frameworks utilize sub-networks or modules that are limited in their ability to capture highly complex relationships in the feature maps and re-weight them accordingly to emphasize the most relevant features.
PAN (Pyramid Attention Network) attempts to address this limitation through the use of a Feature Pyramid Attention (FPA) network module that utilizes convolutions with multiple kernel sizes to perform a spatial attention operation on the feature maps produced by the encoder~\citep{liPyramidAttentionNetwork2018}.
Like the ASPP module in DeepLabV3+, PAN uses this FPA module with the goal of refining large-scale context in the final feature maps produced by the encoder.
However, PAN's FPA module aims to retain localization information in the feature maps through the use of convolutional attention rather than pooling or atrous convolutions.
Similar to Attention U-Net, Global Attention Upsample (GAU) modules are used in the decoder to serve as a form of gating mechanism that attends to the most relevant channels in the feature maps before adding them to the upsampled feature maps from the previous stage of the decoder (as opposed to attending spatially and concatenating the feature maps in Attention U-Net).
ResNet-101 is used as the encoder in the original PAN architecture, and as such no modifications were necessary to incorporate our pre-trained ResNet-101 encoder.

Each model was fine-tuned using the same configuration as the models in Section~\ref{sec:methodology:self_supervised_pretraining} (i.e., AdamW optimizer, learning rate of \(1 \times 10^{-4}\), batch size of 32, focal loss function, AMP, etc.).
During fine-tuning, the weights of the ResNet-101 encoder were unfrozen and updated along with the weights of the semantic segmentation architecture.
Evaluation was again performed using cross-validation across 250, 500, and 750 training samples (Figure~\ref{fig:fold_splits}), and the performance of each model was assessed using each model's macro F1 score on the point-based assessment dataset (25,000 samples) to select the best-performing combination of pre-training procedure, semantic segmentation architecture, and training dataset size using the mean macro F1 score across the k-folds.

\subsection{Model ensembling, large-scale land cover mapping, and validation}\label{sec:methodology:model_ensembling}

Once the best-performing combination of pre-training procedure, semantic segmentation architecture, and training dataset size was determined, we used the models trained using this configuration across all folds to produce a model ensemble for large-scale land cover mapping over the state of Mississippi.
All models in the ensemble shared the same architecture, pre-training procedure, and training dataset size, differing only in the specific samples used for training due to the k-fold cross-validation procedure.
For example, if the best-performing configuration is found to be a U-Net architecture with BYOL pre-training using 750 training samples, then the final ensemble would consist of 4 U-Net models, each pre-trained using BYOL and fine-tuned using 750 training samples from each fold of the cross-validation procedure, with each model having seen a different subset of the training data.
Model ensembling was chosen as it is a popular approach to improving the robustness and accuracy of machine learning models by combining the predictions of multiple distinct models to produce a final prediction.
Each model in the ensemble produced a land cover probability vector for each pixel in the input image, and these class probability vectors were averaged across all models in the ensemble to produce a final class probability vector for each pixel, with the final predicted class being the class with the highest probability from the averaged class probabilities~\citep{marmanisSemanticSegmentationAerial2016}.
A ``sliding window'' approach was used to process the imagery in \SIadj{256}{\meter} patches with a stride of \SIadj{64}{\meter} to ensure that each pixel would be seen multiple times to further maximize the effects of the model ensemble.
Additionally, test-time augmentations were employed to further improve the robustness of the predictions: at any given location, each model in the ensemble received an input that was randomly augmented using random horizontal and vertical flips along with random 90\(^\circ\) rotation~\citep{moshkovTesttimeAugmentationDeep2020}.
For example, if there are 4 models in the ensemble, each model would receive a different random augmentation of the input patch, resulting in 4 different predictions for that patch.
The final output patches of the ensemble were weighted using element-wise multiplication with a 256\(\times\)256 Gaussian kernel to de-emphasize the probabilities of pixels near the edge of the classification maps in order to mitigate presence of edge artifacts in the final merged output.
These inference steps were used to classify the entire state of Mississippi using 2023 NAIP imagery at \SIadj{1}{\meter} resolution, amounting to over 123 billion pixels.
A thematic accuracy assessment of the final land cover product was performed using the point-based assessment dataset (see Section~\ref{sec:methodology:dataset_curation}).
Overall accuracy, user's accuracy, producer's accuracy, F1 Score, Intersection-over-Union (IoU), and Cohen's Kappa (\(\kappa\)) metrics were computed to assess the quality of the final product, as shown in Equations \ref{eq:overall_accuracy}--\ref{eq:kappa}, where TP is the number of true positives, FP is the number of false positives, and FN is the number of false negatives~\citep{congaltonReviewAssessingAccuracy1991,mchughInterraterReliabilityKappa2012}.

\begin{equation}\label{eq:overall_accuracy}
    \text{Overall Accuracy} = \frac{\text{TP} + \text{TN}}{\text{TP} + \text{TN} + \text{FP} + \text{FN}}
\end{equation}

\begin{equation}\label{eq:producer_accuracy}
    \text{Producer's Accuracy} = \frac{\text{TP}}{\text{TP} + \text{FN}}
\end{equation}

\begin{equation}\label{eq:user_accuracy}
    \text{User's Accuracy} = \frac{\text{TP}}{\text{TP} + \text{FP}}
\end{equation}

\begin{equation}\label{eq:f1_score}
    \text{F1 Score} = \frac{2\text{TP}}{2\text{TP} + \text{FP} + \text{FN}}
\end{equation}

\begin{equation}\label{eq:kappa}
    \text{Cohen's } \kappa = \frac{2\times(\text{TP} \times \text{TN} - \text{FN} \times \text{FP})}{(\text{TP} + \text{FP}) \times (\text{FP} + \text{TN}) + (\text{TP} + \text{FN}) \times (\text{FN} + \text{TN})}
\end{equation}

\subsection{Cross-year generalization and validation}\label{sec:methodology:generalization}

To examine the capacity of the model to generalize to other years for the purpose of creating multi-temporal land cover/land cover change products, we applied the final model ensemble to 2016 NAIP imagery over Mississippi to create a secondary land cover product using the same protocol outlined in Section~\ref{sec:methodology:model_ensembling}.
Beyond the creation of a multi-temporal land cover data product, this also enabled an evaluation of the model's capability of handling changes in land cover distribution.
Additionally, NAIP acquisition characteristics vary between years, in terms of both sensor specifications and acquisition period, making it a suitable tool for evaluating the model's response to domain shifts.
As NAIP is a basic aerial imagery product (and does not provide calibrated reflectance values), we adjusted the 2016 imagery to match the spectral characteristics of the 2023 imagery using histogram normalization.
The 2016 product was validated using the same locations as the point-based assessment dataset for 2023 (25,000 samples), but with new human-annotated ground truth labels that correspond to 2016 land cover.

\section{Results}\label{sec:results}

\begin{figure}[ht]
    \centering
    \includegraphics[width=\linewidth]{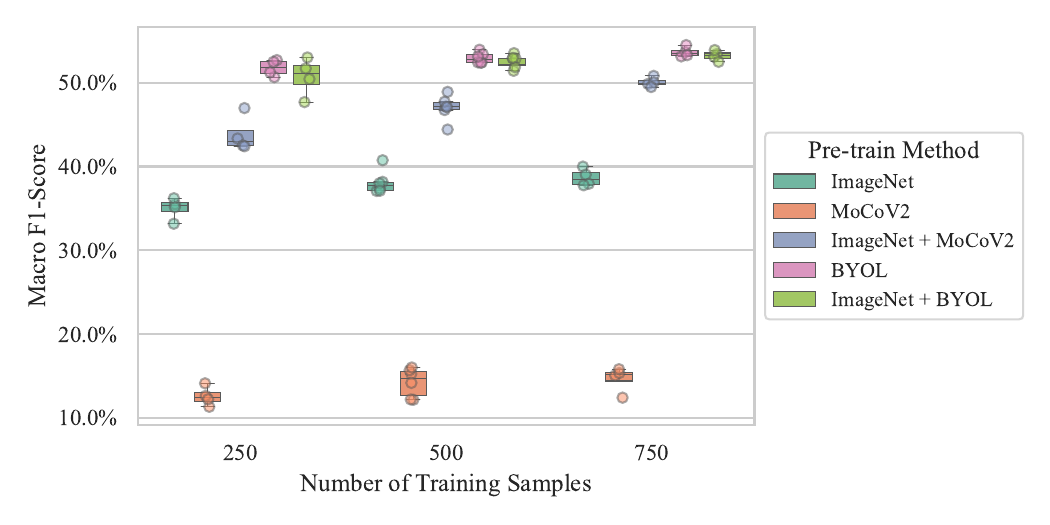}
    \caption{Comparison of linear probing performance using features extracted by ResNet-101 backbones pre-trained with ImageNet only (teal), MoCoV2 only (orange), ImageNet + MoCoV2 (blue), BYOL only (pink), and ImageNet + BYOL (green) given various amounts of labeled training data.}\label{fig:probing_metrics}
\end{figure}

The evaluation of the self-supervised learning approach is divided into two parts: linear probing performance and fine-tuning results.
The linear probing experiments are intended to provide a measure of the quality of the learned image representations produced by the pre-trained encoder in isolation (i.e., without any deep segmentation heads or fine-tuning).
As shown in Figure~\ref{fig:probing_metrics}, using the embeddings from a ResNet-101 model pre-trained using BYOL only as input to a linear classifier yielded an average 15.44\% improvement in macro F1 score over a classifier that used embeddings from a model pre-trained only on ImageNet, which outperformed the models pre-trained only with MoCoV2 in terms of macro F1 score by 23.48\% on average.
This indicates that the BYOL successfully enabled a model to learn to extract features that are linearly separable with regard to land cover classification without the need for ImageNet initialization, while MoCoV2 failed to do so.

Initializing the ResNet-101 model with ImageNet weights prior to MoCoV2 pre-training yielded moderately better performance, resulting in a 9.59\% average increase in macro F1 score compared to simply starting from ImageNet alone, but the BYOL models with ImageNet pre-training outperformed these MoCoV2 models by 5.27\%.
The smallest training dataset size of \(N_\text{train}=250\) exhibited the largest improvement in macro F1 score over ImageNet initialization from BYOL pre-training, with the BYOL pre-trained models achieving a mean macro F1 score of 51.78\% compared to 35.01\% for the ImageNet pre-trained model, an improvement of 16.77\%.
Notably, the embedding model where BYOL pre-training was initialized with ImageNet weights performed marginally worse than the model where BYOL pre-training was performed with random initialization: the average increase in macro F1 score over the ImageNet pre-trained model was 15.44\% for the BYOL model with random initialization, compared to 14.87\% for the BYOL model with ImageNet initialization, a 0.58\% difference.
Overall, the BYOL models showed clear superiority over the MoCoV2 models and the ImageNet model (regardless of initialization strategy), while the MoCoV2 initialized with ImageNet weights showed moderate feature extraction capability compared to an ImageNet model.

\begin{figure*}
    \centering
    \includegraphics[width=\textwidth]{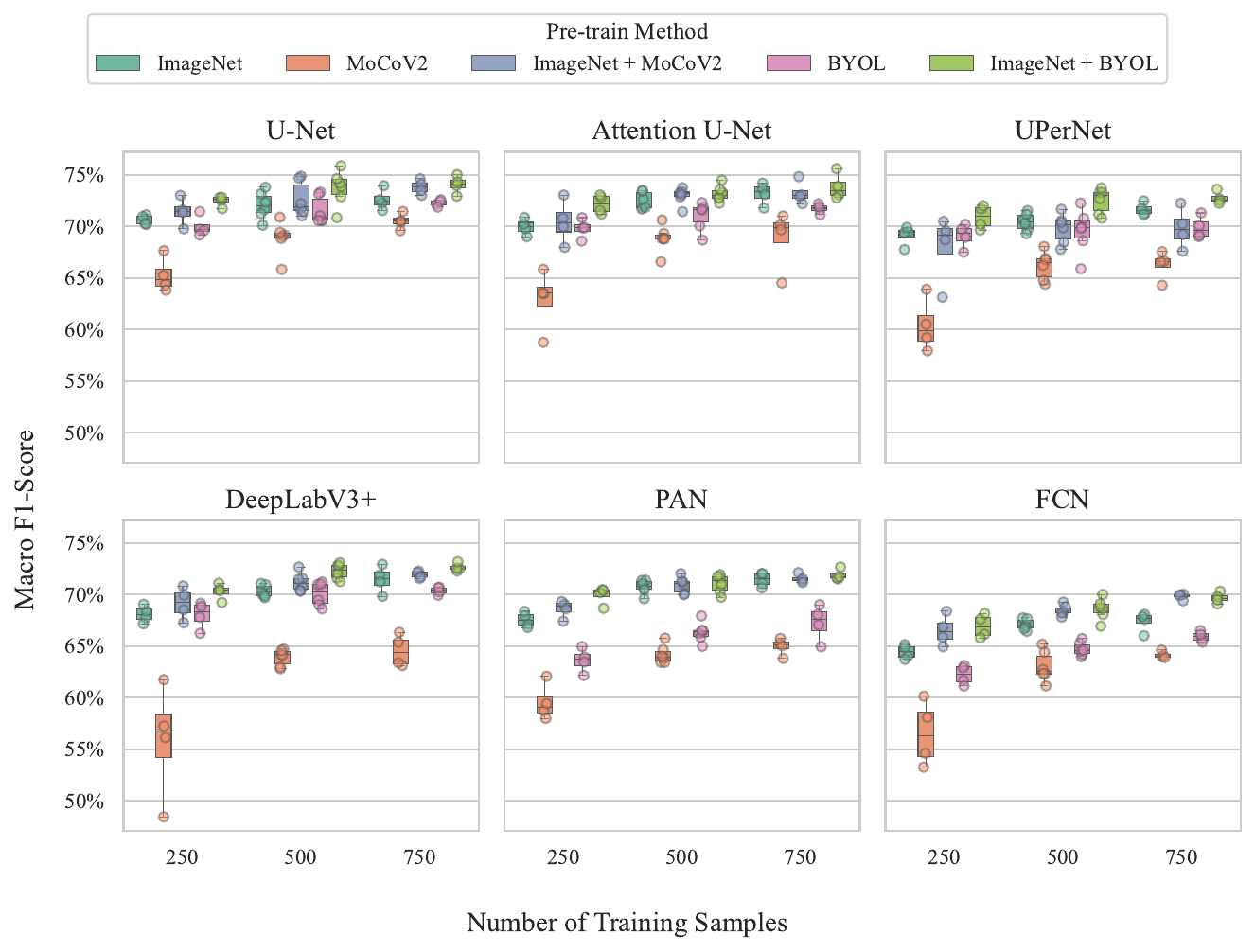}
    \caption{Macro F1 score of semantic segmentation architectures using backbones pre-trained with ImageNet only (teal), MoCoV2 only (orange), ImageNet + MoCoV2 (blue), BYOL only (pink), and ImageNet + BYOL (green) given various amounts of labeled training data.}\label{fig:finetune_metrics}
\end{figure*}

While the linear probing results show vast improvements in representation quality when using BYOL pre-training, fine-tuning these pre-trained encoders using various semantic segmentation architectures resulted in different performance trends.
Figure~\ref{fig:finetune_metrics} shows that the ImageNet encoder outperformed both the BYOL and MoCoV2 pre-trained encoders with random initialization across all combinations of segmentation architectures, with ImageNet models averaging 70.10\% in terms of macro F1 score, a 1.53\% improvement over BYOL and a 5.70\% over MoCoV2.
These results show that, while not achieving the same performance as ImageNet fully supervised pre-training, utilizing self-supervised pre-training with BYOL on domain specific data with randomly initialized weights can be nearly as effective.
Additionally, BYOL with random initialization consistently outperformed MoCoV2 with random initialization across all experimental axes, with an average improvement in macro F1 score of 4.17\%.
MoCoV2 with random initialization also exhibited large sensitivity to shifting training data during cross-validation with an overall standard deviation of 3.39\%, more than double that of BYOL with random initialization (1.52\%) or ImageNet (1.48\%).

Initializing models with ImageNet weights prior to pre-training proved to yield the highest performance: our BYOL and MoCoV2 models pre-trained using this approach achieved average macro F1 scores of 71.63\% and 70.62\%, respectively, increases of 1.53\% and 0.52\% over the ImageNet baseline.
The BYOL models initialized with ImageNet prior to pre-training outperformed all ImageNet models across all test cases, while MoCoV2 outperformed the ImageNet encoder across all cases with the exception of the UPerNet networks.
BYOL with ImageNet pre-training also outperformed MoCoV2 pre-training in all but one case, where the FCN model trained with 750 training samples outperformed its BYOL counterpart by 0.16\%.
BYOL with ImageNet initialization also demonstrated the least variation across all encoder initialization frameworks, with a standard deviation of 1.22\% compared to ImageNet's 1.48\% and ImageNet followed by MoCoV2 pre-training's 1.66\%.

Increasing the training dataset size generally led to improvements in macro F1 score across all architectures and pre-training strategies, though the magnitude of these improvements diminished as the dataset size increased.
For example, the average increase in macro F1 score when increasing the training dataset size from \(N_\text{train}=250\) to \(N_\text{train}=500\) was 2.62\%, while the average increase from \(N_\text{train}=500\) to \(N_\text{train}=750\) was only 0.69\%.
The impact of increasing training dataset size was most pronounced for the MoCoV2 pre-trained models with random initialization, which saw a mean increase of 6.45\% in macro F1 score when increasing the training dataset size from \(N_\text{train}=250\) to \(N_\text{train}=750\), followed by 2.98\% for the ImageNet models, 2.68\% for the MoCoV2 models with ImageNet initialization, 2.45\% for the BYOL models with random initialization, and 2.02\% for the BYOL models with ImageNet initialization.
In terms of semantic segmentation architecture, the U-Net models had the highest mean macro F1 score across all training dataset sizes and pre-training strategies (71.48\%), followed by Attention U-Net (71.06\%), UPerNet (69.16\%), DeepLabV3+ (68.80\%), PAN (68.07\%), and FCN (65.79\%).
Weak positive correlation between model size (in terms of number of parameters) and macro F1 score was observed (\(r=0.30\)). A comprehensive table of performance metrics across all models tested is provided in Table S1 in the supplementary materials.

\begin{figure*}
    \centering
    \includegraphics[width=\textwidth]{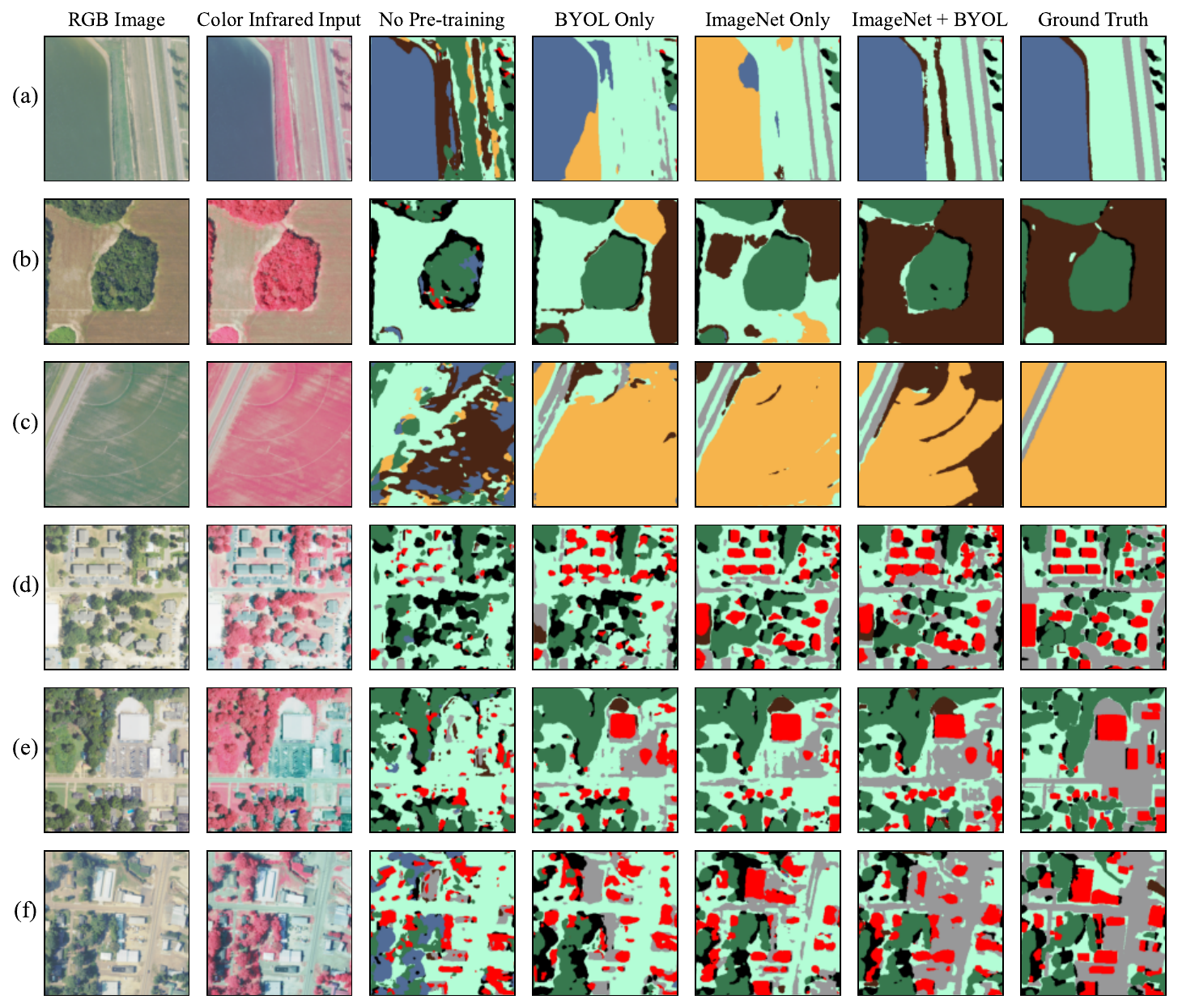}
    \caption{Visual comparison of land cover classification results from a U-Net model with \(N_\text{train} = 500\) using k-fold run \protect\#2.
    }\label{fig:visual_assessment}
\end{figure*}

The best-performing pre-training, semantic segmentation architecture, and training dataset size combination based on mean macro F1 score across all k-folds was the BYOL model initialized with ImageNet weights used as the backbone in a U-Net architecture trained with \(N_\text{train}=750\) samples, which achieved a mean macro F1 score of 74.04\% and a mean overall accuracy of 86.43\%, a 1.47\% and 0.32\% improvement over the U-Net model using the same training dataset size but with an ImageNet pre-trained backbone, respectively.
To illustrate the qualitative benefits of the BYOL pre-training approach, we highlight a particularly strong individual model run. Run \#2 of the U-Net model with the BYOL backbone (ImageNet initialization) fine-tuned using \(N_\text{train}=500\) samples achieved a macro F1 score of 75.84\%.
A qualitative assessment of this model's inference results is shown in Figure~\ref{fig:visual_assessment}, demonstrating that not only did the model outperform its ImageNet-only counterpart quantitatively (an increase of 2.06\% macro F1 score for the same data split), but it also produced land cover maps with sharper object boundaries and fewer classification artifacts.
Samples (a), (b), and (c) in Figure~\ref{fig:visual_assessment} show overall improvements in classification accuracy with additional pre-training across multiple land cover classes, particularly in homogeneous regions such as water bodies and fields (though some misclassifications still occur, such as in (c)).
The importance of object localization is highlighted in samples (d), (e), and (f), where the ImageNet-only model tends to yield acceptable overall classification accuracy, but fails to accurately delineate small features such as roads and buildings, which are better captured by the model with additional BYOL pre-training.
While these examples are illustrative and do not change the overall ranking reported above, they highlight the qualitative improvements of the BYOL pre-training approach, which are particularly important given the \SIadj{1}{\meter} spatial resolution of the ground truth labels.

\begin{figure*}
    \centering
    \includegraphics[width=\textwidth]{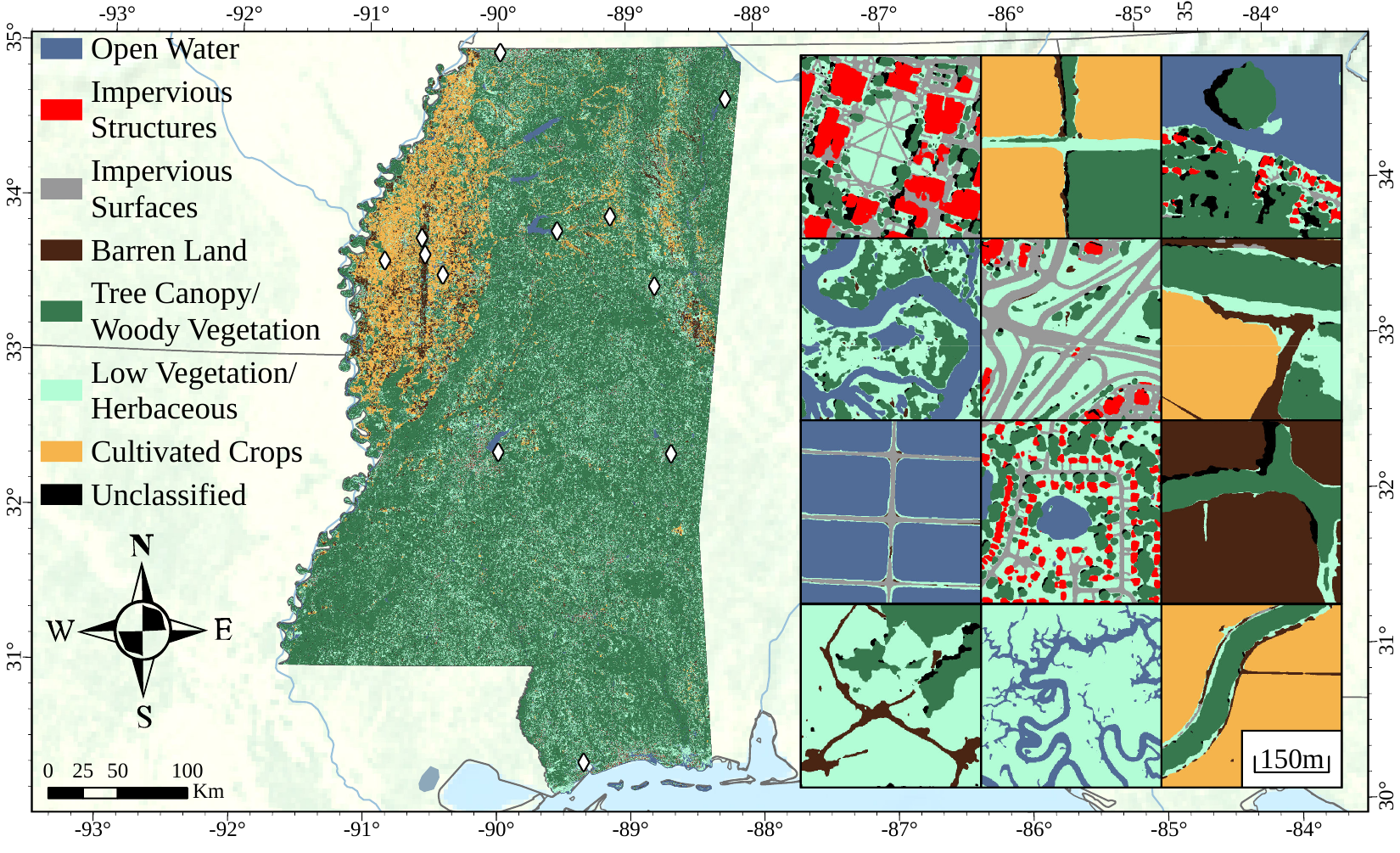}
    \caption{Statewide 2023 land cover inference results for Mississippi (left) and selected zoomed-in regions (right).}\label{fig:inference_results}
\end{figure*}

Despite this specific model's strong individual performance, it was an outlier in terms of performance across all k-folds, as the mean macro F1 score and overall accuracy for this model were 73.70\% and 86.55\%, respectively, falling short of the best-performing configuration in terms of mean macro F1 score.
As such, the U-Net model with a BYOL backbone initialized with ImageNet weights and fine-tuned using \(N_\text{train}=750\) samples was considered the best-performing model overall and was used to generate the final land cover map for the entire state of Mississippi.
The 4 models trained using this best-performing configuration as part of the k-fold cross-validation procedure were subsequently used to generate a final land cover map for the entire state of Mississippi using the model ensembling and test-time augmentation procedure outlined in Section~\ref{sec:methodology:model_ensembling}.
A visualization of the final comprehensive \SIadj{1}{\meter} resolution land cover map for Mississippi is shown in Figure~\ref{fig:inference_results}.
We note the presence of a large vertical stripe of barren land in the Mississippi Delta region that stands in contrast to the surrounding cropland; this is an artifact of a data acquisition anomaly in the 2023 NAIP imagery, where a portion of the imagery was collected during a period when fields were fallow and thus appeared barren.
This highlights a unique challenge of mapping agricultural activity using imagery acquired across multiple dates, as phenological changes in cropland can cause inconsistent land cover classifications.
Still, despite this anomaly, our final land cover product offers a complete and precise spatial inventory of natural resources and human developments across the entire state.
For example, water bodies appear to be delineated with high precision, revealing the presence of small streams and ponds, forested areas are well captured with fragmentation appropriate to the landscape context (e.g., trees interspersed within urban developments), and developed areas with small buildings and road networks are distinct and mapped with adequate spatial detail.

The confusion matrix from final accuracy assessment is shown in Table \ref{tab:confusion_matrix}, while Table \ref{tab:class_wise_metrics} gives a comprehensive breakdown of the model's per-class and overall accuracy metrics, demonstrating that our model yielded high accuracy when classifying natural resources, but struggled with land cover types that are more anthropogenically influenced.
For example, the barren land class had a moderately high user's accuracy of 82.66\%, but a low producer's accuracy of 57.42\%, and the model had a tendency to misclassify barren land pixels as impervious surfaces, herbaceous, and cultivated crops.
The cultivated crops class suffered from the opposite problem, with a high producer's accuracy of 95.54\% but a low user's accuracy of 65.91\%, with barren land and herbaceous classes being frequently misclassified as cultivated crops.
On the other hand, the open water and tree canopy/woody vegetation classes were classified with high accuracy, with F1 scores of 92.91\% and 94.93\%, respectively, highlighting the model's strength in identifying these natural features.
Overall, the final product's macro F1 score and overall accuracy were 75.58\% and 87.14\%, respectively, while the mean class-wise user's and producer's accuracy were 80.18\% and 75.60\%, respectively.
We do note that the large amount of forested land in Mississippi makes a balanced assessment challenging, as forested land accounted for over 60\% of all sampled assessment points, making the overall accuracy metric less informative due to the class imbalance: we thus emphasize the macro F1 score and individual class-wise metrics as a more holistic measure of model performance in this context.
A thorough quantitative and qualitative assessment of errors is given in Section~\ref{sec:sup:error_assessment} of the supplementary materials.

\begin{table*}
  \centering
  \scriptsize
  \caption{Confusion matrix for the 2023 Mississippi land cover classification results.}\label{tab:confusion_matrix}
  \begin{tabular*}{\textwidth}{@{\extracolsep{\fill}}lrrrrrrrrr}
    \toprule
    & \multicolumn{9}{l}{\textbf{Predicted Class}} \\
    \cmidrule(lr){2-10}
    \textbf{True Class} & \bfseries \specialcell[t]{Open\\Water} & \bfseries \specialcell[t]{Impervious\\Structures} & \bfseries \specialcell[t]{Impervious\\Surfaces} & \bfseries \specialcell[t]{Barren\\Land} & \bfseries \specialcell[t]{Tree Canopy/Woody\\Vegetation} & \bfseries \specialcell[t]{Low Vegetation/\\Herbaceous} & \bfseries \specialcell[t]{Cultivated\\Crops} & \bfseries Unclassified & \bfseries Total \\
    \midrule
    Open Water            & \textbf{511} & 1 & 0 & 10 & 11 & 18 & 0 & 21 & 572 \\
    Impervious Structures & 0 & \textbf{80} & 15 & 3 & 3 & 5 & 0 & 0 & 106 \\
    Impervious Surfaces   & 1 & 5 & \textbf{183} & 17 & 1 & 22 & 0 & 1 & 230 \\
    Barren Land           & 6 & 4 & 68 & \textbf{944} & 22 & 451 & 144 & 5 & 1,644 \\
    Tree Canopy           & 1 & 2 & 2 & 6 & \textbf{14,712} & 366 & 6 & 22 & 15,117 \\
    Low Vegetation        & 7 & 2 & 20 & 138 & 376 & \textbf{3,699} & 492 & 17 & 4,751 \\
    Cultivated Crops      & 0 & 0 & 0 & 17 & 2 & 38 & \textbf{1,243} & 1 & 1,301 \\
    Unclassified          & 2 & 11 & 6 & 7 & 753 & 86 & 1 & \textbf{413} & 1,279 \\
    \midrule
    Total                 & 528 & 105 & 294 & 1,142 & 15,880 & 4,685 & 1,886 & 480 & \textbf{25,000} \\
    \bottomrule
  \end{tabular*}
\end{table*}

\begin{table*}
  \centering
  \caption{Per-class classification metrics for the 2023 Mississippi land cover classification results.}\label{tab:class_wise_metrics}
  \begin{tabular*}{\linewidth}{@{\extracolsep{\fill}}llllll}
    \toprule
    \bfseries Class Name & \bfseries User's Accuracy (\%) & \bfseries Producer's Accuracy (\%) & \bfseries F1 Score (\%) & \bfseries IoU (\%) & \bfseries Cohen's \(\kappa\) \\
    \midrule
    Open Water                & 96.78 & 89.34 & 92.91 & 86.76 & 0.9275 \\
    Impervious Structures     & 76.19 & 75.47 & 75.83 & 61.07 & 0.7573 \\
    Impervious Surfaces       & 62.24 & 79.57 & 69.85 & 53.67 & 0.6953 \\
    Barren Land               & 82.66 & 57.42 & 67.77 & 51.25 & 0.6593 \\
    Tree Canopy / Woody Veg.  & 92.64 & 97.32 & 94.93 & 90.34 & 0.8666 \\
    Low Veg. / Herbaceous     & 78.95 & 77.86 & 78.40 & 64.48 & 0.7338 \\
    Cultivated Crops          & 65.91 & 95.54 & 78.00 & 63.94 & 0.7656 \\
    Unclassified              & 86.04 & 32.29 & 46.96 & 30.68 & 0.4543 \\
    \midrule
    Macro Average             & 80.18 & 75.60 & 75.58 & 62.77 & 0.7325 \\
    Overall                   & Accuracy: 87.14 & & & 77.21 & 0.7751 \\
    \bottomrule
  \end{tabular*}
\end{table*}

\begin{figure*}
    \centering
    \includegraphics[width=\textwidth]{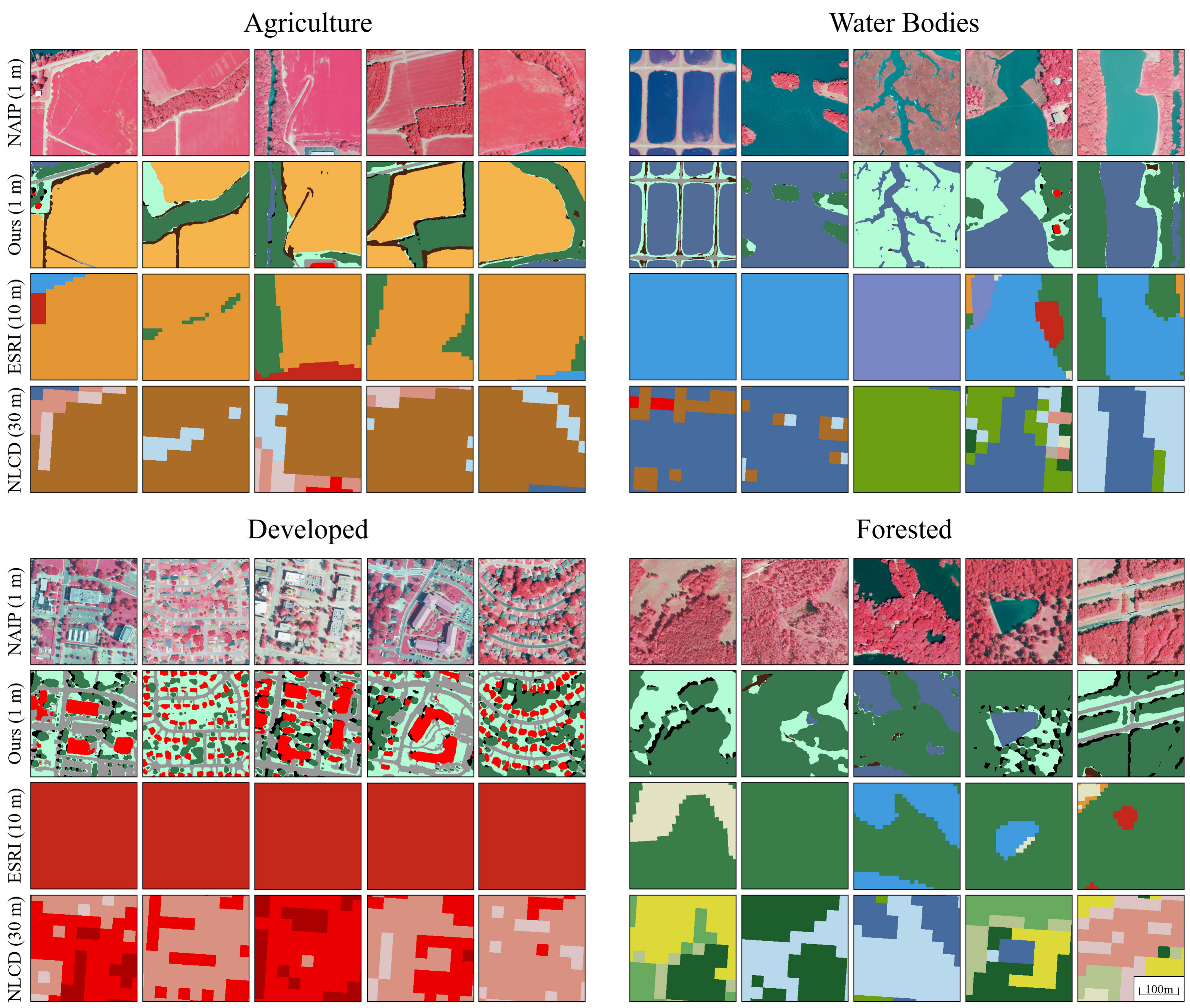}
    \caption{Comparison of \SIadj{1}{\meter} resolution input imagery from 2023, our \SIadj{1}{\meter} resolution land cover classification results, \SIadj{10}{\meter} land cover data from the ESRI annual land use/land cover dataset, and \SIadj{30}{\meter} land cover data from NLCD over various land cover types in Mississippi, USA.}\label{fig:lc_comparison}
\end{figure*}

Figure~\ref{fig:lc_comparison} provides a visual comparison of the final land cover product against comparable existing land cover datasets (NLCD, ESRI Annual Land Use Land Cover~\citep{karraGlobalLandUse2021}) over various land cover settings in Mississippi, illustrating the increase in spatial detail our \SIadj{1}{\meter} resolution product provides over existing products with coarser spatial resolutions (\SIadj{10}{\meter} for ESRI LULC and \SIadj{30}{\meter} for NLCD).
We observe that our product is more suitable for applications requiring fine spatial detail, high spatial accuracy, and precise object boundaries given the improved spatial resolution over these existing products.
For example, agricultural fields are precisely delineated in our product, while they appear as coarse blocks of land cover in the NLCD and ESRI LULC products.
In urban areas, the mix of natural and developed land cover types interspersed within residential neighborhoods is well captured in our product, while the coarser products tend to generalize these areas into a single land cover type.
In forested areas, small clearings, roads, and barren patches are captured in our product, while they are often absent in the coarser products.
However, these existing products are derived from satellite imagery composites and represent the land cover over the span of a given year, while our product is derived from aerial imagery collected during a specific acquisition window in 2023.
As observed in the Mississippi Delta region, this can lead to discrepancies in land cover classification due to phenological changes in agricultural land cover over time, which is a limitation of our product compared to these existing products that integrate data over longer time periods.

The final land cover product's macro F1 score represents an increase of 1.54\% over the mean performance of the base k-fold models, which can be attributed to the model ensembling and test-time augmentation procedures outlined in Section~\ref{sec:methodology:model_ensembling}.
Table~\ref{tab:final_f1_comparison} provides a class-wise comparison of the F1 scores between the best overall performing model, the mean performance of the models in the ensemble, the best performing model in the ensemble, and the model ensemble itself.
Notably, the impervious surfaces and barren land classes saw dramatic improvements in F1 score compared to the mean performance of the models in the ensemble, with increases of 6.46\% and 3.81\%, respectively.
Other classes saw more modest improvements, while the unclassified class showed a decrease of 2.09\% in F1 score after ensembling.
The best performing model in the ensemble (from cross-validation run \#2) outperformed the ensemble itself when classifying impervious structures (+1.09\% F1 score), but was more than offset by the ensemble's capacity for classifying impervious surfaces (+5.51\% F1 score), leading to an improvement of 0.58\% macro F1 score and 0.53\% overall accuracy.

\begin{table*}
    \centering
    \caption{Comparison between mean F1 scores from the best-performing overall individual model (U-Net, ImageNet + BYOL backbone, \(N_\text{train}=500\), cross-validation run \#2), best performing k-fold model configuration (U-Net with BYOL backbone initialized with ImageNet weights and fine-tuned using \(N_\text{train}=750\) samples), best performing individual model within the ensemble (run \#2), and the final F1 scores from the ensemble model used to generate the final land cover product.}\label{tab:final_f1_comparison}
    \scriptsize
    \begin{tabular*}{\linewidth}{@{\extracolsep{\fill}}lllllllllll}
    \toprule
    & \bfseries \specialcell[t]{Open\\Water} & \bfseries \specialcell[t]{Impervious\\Structures} & \bfseries \specialcell[t]{Impervious\\Surfaces} & \bfseries \specialcell[t]{Barren\\Land} & \bfseries \specialcell[t]{Tree Canopy/\\Woody Vegetation} & \bfseries \specialcell[t]{Low Vegetation/\\Herbaceous} & \bfseries \specialcell[t]{Cultivated\\Crops} & \bfseries \specialcell[t]{Unclassified} & \bfseries \specialcell[t]{Macro\\Average} & \bfseries Overall \\
    \midrule
    \specialcell{Best model overall \\ F1 Score (\%)} & 92.74 & 75.98 & 62.74 & \bfseries 68.10 & \bfseries 95.12 & 78.11 & \bfseries 81.23 & \bfseries 52.73 & \bfseries 75.84 & \bfseries 87.42 \\
    \addlinespace
    \specialcell{Mean Ensemble \\ F1 Score (\%)} & 92.24 & 75.28 & 63.39 & 63.96 & 94.84 & 77.01 & 76.55 & 49.04 & 74.04 & 86.43 \\
    \addlinespace
    \specialcell{Best model in ensemble \\ F1 Score (\%)} & 92.66 & \bfseries 76.92 & 64.34 & 67.07 & 94.88 & 76.92 & 76.34 & 50.84 & 75.00 & 86.61 \\
    \addlinespace
    \specialcell{Final F1 Score (\%)} & \bfseries 92.91 & 75.83 & \bfseries 69.85 & 67.77 & 94.93 & \bfseries 78.40 & \bfseries 78.00 & 46.96 & 75.58 & 87.14 \\
    \bottomrule
    \end{tabular*}
\end{table*}

When compared to the best performing model overall (U-Net, ImageNet + BYOL backbone, \(N_\text{train}=500\), cross-validation run \#2), we observe that the ensemble performed similarly on aggregate, with a marginal 0.26\% decrease in macro F1 score and a 0.28\% decrease in overall accuracy, but the ensemble yielded a 7.11\% increase in F1 score in the classification of impervious surfaces, a critical minority class for VHSR land cover data.
The best individual model's primary advantage over the ensemble was the classification of “unclassified” regions, with an improvement of 5.77\% in terms of F1 score.
However, as the unclassified class provides little utility for downstream analysis, we selected the ensemble model for its improved accuracy in the impervious surface class for robust mapping of urban regions.
The use of the model ensemble approximately quadrupled the time needed for inference compared to a single model, highlighting the trade-off between accuracy and computational efficiency when using model ensembling and test-time augmentation.
For our purposes, we deemed the increase in accuracy worth the additional computational cost, which was not prohibitive given the available computational resources, and elected to use the ensembled predictions for the final land cover product.

\begin{figure*}
    \centering
    \includegraphics[width=\textwidth]{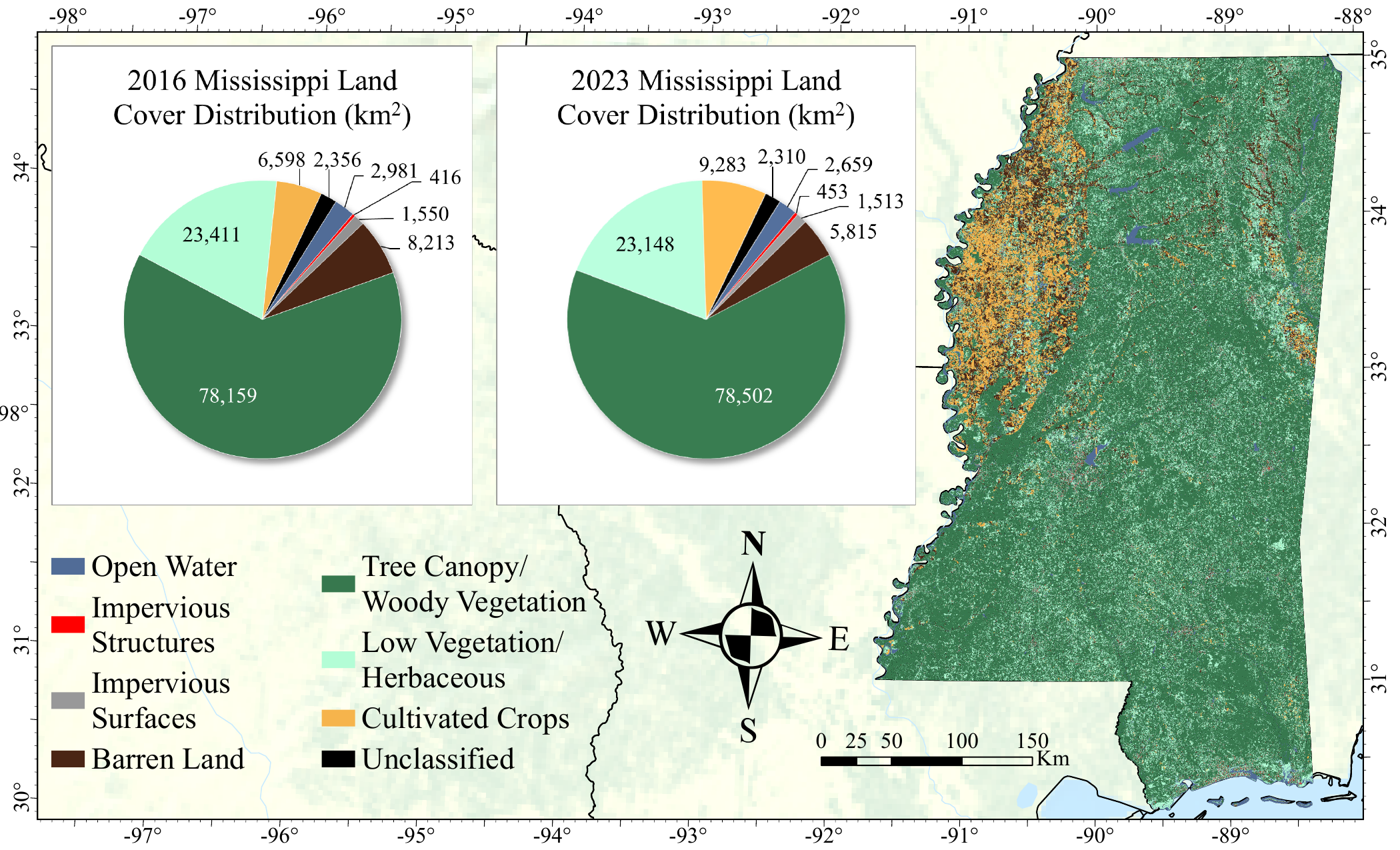}
    \caption{Comparison of land cover distribution between 2016 and 2023 (left) and visualization of 2016 land cover data (right).}\label{fig:inference_2016}
\end{figure*}

As shown in Table~\ref{tab:comparison_2016}, the model was able to classify 2016 imagery well despite not being trained on this data. Only relatively small decreases of 2.36\% and 1.93\% were observed in overall accuracy and macro F1 score.
However, there was considerable variation in class-wise performance between the two years: for example, the 2016 product yielded a 5.73\% improvement in F1 score over the 2023 product when classifying barren land classes, but showed a 9.05\% decrease in F1 score when classifying cultivated crops.
One potential cause for drastic shift in classification accuracy is the difference in acquisition schedule for the 2016 imagery compared to the 2023 imagery.
Most of the imagery for 2016 was acquired in early June (either during or shortly after the planting period for soybeans~\citep{sunEvaluationModelsSimulating2022}) whereas a majority of the 2023 imagery was acquired in mid-to-late August.
Figure~\ref{fig:inference_2016} shows that the model predicted 28.92\% less cropland in 2016 compared to 2023, while the predicted barren land area increased by 41.24\%.
Examining the spatial distribution of the 2016 land cover data shows that most of the larger swaths of barren land lie in the Mississippi Delta region of the state interleaved with other cultivated crops.
This highlights the unique challenge of mapping agricultural activity from mono-temporal aerial imagery, particularly in Mississippi where a wide range of crops with varying phenological characteristics are grown, making it difficult to capture agricultural activity comprehensively.
Still, the model performed similarly on the 2016 data as the 2023 data across all other classes, with the exception of the unclassified class.
We provide a confusion matrix in Table~\ref{tab:confusion_matrix_2016} of the supplementary materials, along with class-wise accuracy metrics in Table~\ref{tab:class_wise_metrics_2016}.

\begin{table*}
    \centering
    \caption{Class-wise comparison of F1 scores of 2023 product and 2016 product.}\label{tab:comparison_2016}
    \scriptsize
    \begin{tabular*}{\linewidth}{@{\extracolsep{\fill}}lllllllllllll}
    \toprule
    & \bfseries \specialcell[t]{Open\\ Water} & \bfseries \specialcell[t]{Impervious\\ Structures} & \bfseries \specialcell[t]{Impervious\\ Surfaces} & \bfseries \specialcell[t]{Barren\\ Land} & \bfseries \specialcell[t]{Tree Canopy/\\ Woody Vegetation} & \bfseries \specialcell[t]{Low Vegetation/\\ Herbaceous} & \bfseries \specialcell[t]{Cultivated\\ Crops} & \bfseries Unclassified & \bfseries \specialcell[t]{Macro\\ Average} & \bfseries Overall \\
    \midrule
    \bfseries 2023 F1 Score (\%) & 92.91 & 75.83 & 69.85 & 67.77 & 94.93 & 78.40 & 78.00 & 46.96 & 75.58 & 87.14 \\
    \addlinespace
    \bfseries 2016 F1 Score (\%) & 90.26 & 76.53 & 71.82 & 73.49 & 93.26 & 77.18 & 68.96 & 37.70 & 73.65 & 84.78 \\
    \bottomrule
    \end{tabular*}
\end{table*}

\section{Discussion}\label{sec:discussion}

This study demonstrated that deep learning semantic segmentation models for dense land cover classification at VHSR can be effectively trained with limited labeled data, and that the resulting models can be used to produce highly detailed land cover maps over large areas.
As shown in Figure~\ref{fig:lc_comparison}, the final land cover product produced using our approach provides a detailed characterization of surface phenomena across the entire state of Mississippi with improved spatial fidelity compared to existing data products such as \SIadj{30}{\meter} NLCD data and \SIadj{10}{\meter} data from the ESRI annual land cover dataset~\citep{karraGlobalLandUse2021}.
Our land cover product captures fine-scale land cover features such as individual buildings, roads, small bodies of water, precise agricultural field delineations, forest clearings, and mixed urban-vegetation interfaces that are indistinguishable in the NLCD land cover data.
As~\citet{boyleHighResolutionSatelliteImagery2014} and~\citet{fisherImpactSatelliteImagery2018} noted, the ability to classify fine-scale land cover features is critical for a wide variety of applications, and has been shown to have a substantial impact on the accuracy and utility of downstream applications that rely on land cover data, such as water quality modeling, land use assessment, and conservation biology.
However, the 8 land cover classes used in our product are more general than those used in the NLCD dataset, which can limit the expressiveness of our product in certain contexts (e.g., distinguishing between different forest types, wetlands, developed intensities, etc.).
Additionally, confusion between certain classes is evident in the thematic accuracy assessment results shown in Table~\ref{tab:confusion_matrix}, demonstrating that while overall accuracy is high, certain classes remain challenging to classify accurately.
For example, there is substantial confusion between the barren land and impervious surfaces classes, which is likely due to the fact that these surfaces are spectrally and texturally similar in the NAIP imagery used in this work.
Still, given the high spatial resolution and novelty of our data product, we believe it will be highly suitable for downstream work that requires highly detailed information of surface features at both small and large scales.

Core to the success of our approach is the use of self-supervised learning to pre-train an image encoder using large amounts of unannotated NAIP imagery prior to fine-tuning on the target land cover classification task.
We achieved high land cover classification accuracy with only 1,000 labeled training samples, which is a small fraction of the amount of labeled data typically used to train deep learning models for similar tasks.
For example, the ISPRS Potsdam and Vaihingen datasets can be used to create 23,750 and 8,873 256\(\times\)256 pixel training patches, respectively~\citep{2DSemanticLabeling,2DSemanticLabel}.
This corresponds to an approximately 96\% and 89\% reduction in the quantity of labeled data required compared to these benchmarks, representing a substantial decrease in the manual resources necessary for annotation.
Additionally, these datasets only cover small urban areas, while our work concerns a substantially larger extent with multiple diverse ecoregions.
The results of our experiment provide strong evidence to support this claim, and we applied this methodology to produce a \SIadj{1}{\meter} resolution land cover map for the entire state of Mississippi.
We note that for linear probing, the embeddings produced by the encoder pre-trained using BYOL alone performed the best, slightly outperforming the BYOL model initialized with ImageNet weights and substantially outperforming the ImageNet pre-trained model.
One observation from the quantitative assessment of the model is the discrepancy between the performance of the pre-trained models when used as fixed feature extractors for linear probing compared to when they are fine-tuned as part of an end-to-end trainable semantic segmentation architecture.
When only working with fixed embeddings, the BYOL model with random initialization outperformed the BYOL model initialized with ImageNet weights, which in turn outperformed the ImageNet pre-trained model.
This suggests that the embeddings produced by the BYOL model with random initialization are somewhat linearly separable with respect to the land cover classes of interest, even in the absence of any supervised fine-tuning.
This is a promising result, as it contributes to the growing body of evidence that self-supervised learning methods can be just as effective (if not more so) than traditional supervised pre-training methods for remote sensing applications~\citep{liGlobalLocalContrastive2022,bergSelfSupervisedLearningScene2022,taoSelfSupervisedRemoteSensing2023}.
These results also provided a quantitative demonstration of the domain gap between the natural, general-purpose images found in ImageNet and remote sensing imagery, and the benefits of pre-training on in-domain data, as echoed by findings in other recent works~\citep{neumannTrainingGeneralRepresentations2020}.
However, when fine-tuning the entire model as part of a semantic segmentation architecture, the BYOL model initialized with ImageNet weights consistently outperformed both the BYOL model with random initialization and the ImageNet pre-trained model across all architectures and training dataset sizes.
This is a critical observation, as it indicates that the broader, more general image features learned from ImageNet are potentially relevant to the land cover classification task when a more complex decoder network is used to learn task-specific features during fine-tuning, suggesting that in spite of the apparent domain gap, ImageNet pre-training is still relevant for remote sensing applications.

We found that across all cases, the BYOL consistently outperformed MoCoV2 in both linear probing and fine-tuning performance.
This is notable as BYOL is ultimately a simpler approach to pre-training compared to MoCoV2, foregoing requirements for large numbers of negative samples required by the contrastive loss function and requiring fewer hyper-parameters to tune.
For example, beyond the standard set of hyper-parameters for model training, MoCoV2 performance tends to vary under different values for momentum (parameter \(m\) in Equation~\ref{eq:ema_update}), temperature (parameter \(\tau\) in Equation~\ref{eq:infonce_loss}), queue size, and projection head embedding dimension, while BYOL only requires tuning of the momentum parameter and projection head embedding dimension.
For all experiments, we used parameters discovered by the original authors of the respective works determined to yield good results; however, given the domain shift between general purpose images and aerial imagery, it is possible that tuning these hyperparameters may yield additional performance.
However, tuning these is computationally expensive due to the large amount of training data required and long training schedules.
Additionally, the contrastive loss may be ill-posed for this task due to the homogeneity of the pre-training dataset. For example, Figure~\ref{fig:pretraining_samples} shows that in the pre-training data, many separate samples share very similar characteristics due to the fact that a large portion of Mississippi's landscape is dominated by forested land cover or cultivated crops.
When using a contrastive loss, the model may be penalized for outputting similar embeddings for negative pairs even when those negative pairs share similar semantic content (i.e., “false negatives”).
For example, even though two tiles consist completely of forested area, they are still treated as negative pairs for the purpose of optimization.
Increasing the temperature parameter to smooth the optimization landscape or reducing the queue size may help alleviate these issues, but this is a well-recognized fundamental flaw in applying contrastive learning to homogeneous datasets~\citep{huynhBoostingContrastiveSelfSupervised2022,chenIncrementalFalseNegative2022,zhangFALSEFalseNegative2022}.
BYOL's self-distillation framework avoids this scenario entirely by incorporating asymmetry in the pre-training in order to give the online encoder a robust training signal without the need for negative samples.

A key feature of the BYOL approach used in this study is that it is model-agnostic in nature, meaning that it can theoretically be applied to any CNN or Transformer-based image encoder architecture, and these encoders can then be transferred into any downstream architecture for fine-tuning.
On the other hand, many approaches to foundation model pre-training for remote sensing applications currently rely on ViT-based architectures~\citep{liPredictingGradientBetter2024,yaoRingMoSenseRemoteSensing2023,dumeurSelfSupervisedSpatioTemporalRepresentation2024}, which have not yet seen wide adoption for dense prediction tasks in remote sensing due to the computational complexity associated with training and inference using such models, while CNN-based semantic segmentation architectures are still widely used and well understood in the remote sensing community~\citep{xuVisionTransformerExcellent2022,zhuChangeViTUnleashingPlain2026}.
Additionally, given recent results in computer vision research that show that CNNs can achieve comparable performance to Transformers on a variety of computer vision tasks when designed appropriately~\citep{liuConvNet2020s2022,wooConvNeXtV2CoDesigning2023}, it is still unclear whether Transformer-based architectures are strictly superior to CNNs for remote sensing applications across the board.
Thus, the model-agnostic nature of BYOL is a significant advantage, as it allows for flexibility in choosing the most appropriate architecture for the target task without being constrained to a specific model type.
For example, we transferred the BYOL pre-trained ResNet-101 encoder into multiple semantic segmentation architectures (FCN, U-Net, Attention U-Net, DeepLabV3+, UPerNet, PAN) and found that using the BYOL pre-trained encoder consistently improved performance across all architectures compared to using an ImageNet pre-trained encoder alone.
Given the wide availability of unannotated VHSR imagery from sources such as NAIP and government satellite programs, we believe that self-supervised learning methods such as BYOL provide a promising avenue for future research in remote sensing applications, where labeled data is often scarce but unannotated data is abundant.

In terms of architectures, we found that the U-Net models tended to outperform all other architectures tested, despite being among the simpler frameworks for semantic segmentation.
One potential reason for this is the use of a symmetric decoder framework with skip connections instead of a shallow decoder network (as used in DeepLabV3+ and FCN).
This can also be observed based on the performance of the Attention U-Net and UPerNet architectures, which follow similar symmetric encoder-decoder architectures.
However, these models integrate complex sub-networks in an attempt to improve performance, such as attention gates and pyramid pooling. While these sub-networks may facilitate better multi-scale feature extraction or spatial feature localization when sufficient data is available for training, under limited data scenarios they appear to hinder the performance of the full networks~\citep{brigatoCloseLookDeep2021}.
On the other hand, the U-Net decoder consists entirely of 3\(\times\)3 convolutions, batch normalization, and bilinear upsampling layers: this has the effect of making it straightforward to fit with few training samples, but reduces the effective receptive field of the network, making incorporating long-range context into classifications difficult.
Section~\ref{sec:sup:error_assessment} in the supplementary materials includes a short discussion on how this impacts the classification of large buildings, leading to confusion between the impervious surfaces and impervious structures classes.

We note that while our results demonstrate the effectiveness of BYOL for pre-training an image encoder for land cover classification at VHSR, other self-supervised learning methods may also be effective for this task.
One notable restriction of BYOL, MoCoV2, and similar transformation-based self-supervised learning methods is that they tend to rely on image augmentation routines that are specifically designed for three-channel RGB imagery, which can limit their applicability to multispectral or hyperspectral remote sensing imagery that contains more than three bands.
For example, a central component of BYOL's augmentation pipeline involves adjusting the brightness, contrast, saturation, and hue of an input image.
The saturation and hue adjustment transformations are explicitly defined for three-channel imagery as a mapping from the RGB color space to the HSV color space, and thus cannot be directly applied to imagery with more than three bands.
For remote sensing applications, this is a significant limitation of transformation-based self-supervised learning methods that rely on image augmentations to learn transformation-invariant image representations, restricting their applicability to only 3-band imagery.
While we achieved strong results using BYOL on 3-band CIR NAIP imagery, 4 or more bands may be necessary to effectively utilize spectral information for other remote sensing applications, such as vegetation mapping, crop classification, or soil property estimation.
In the case of land cover classification, we found that CIR composites are sufficient for robust land cover classification tasks, especially given that spectral reflectance in the blue band is unreliable due to atmospheric scattering~\citep{royConterminousUnitedStates2014,juContinentalscaleValidationMODISbased2012,vermoteAtmosphericCorrectionMonitoring2008}, and historically has provided little utility for other land cover/land use classification tasks~\citep{aminiUrbanLandUse2022,temenosInterpretableDeepLearning2023}.
Still, our model struggles to differentiate between impervious surfaces and barren land (see Table~\ref{tab:confusion_matrix} and sample (b) in Figure~\ref{fig:error_analysis} in the supplementary materials), and given the utility of the blue band for soil mapping~\citep{liuNDBSINormalizedDifference2022,royTropicalForestCover2002}, the inclusion of this band could increase the spectral separability of the two classes.
Future work should continue to explore the implications of self-supervised learning for multispectral remote sensing applications under real-world constraints beyond academic benchmarks and compute-intensive techniques.

Further, for label-scarce scenarios, little information on optimal sampling strategies for patch-based dense classification tasks is currently available in the literature, as opposed to more well-studied pixel-based sampling strategies for traditional machine learning classifiers~\citep{brownLessonsLearnedImplementing2020,zhuOptimizingSelectionTraining2016,stehmanImpactSampleSize2012}.
Our approach relied on using existing coarse-resolution land cover data to select a representative set of training samples, but this approach may introduce bias into the training dataset given that the NLCD data is derived from another classification algorithm, not a human annotator, and the resolution is too coarse to capture fine details in land cover, such as small roads or barren clearings (see Figure~\ref{fig:lc_comparison}).
Subsequently, the feature vectors derived from this data and used for stratification may not be representative of high-frequency spatial land cover patterns, introducing bias into the dataset.
Examining the ratio between number of predicted samples and number of actual samples for each class in Table~\ref{tab:confusion_matrix} gives an insight into specific classes that our model is biased towards or against.
For example, in the 2023 assessment, the model predicted 1,886 pixels as cultivated crops, while the actual number of pixels corresponding cultivated crops is only 1,301.
On the other hand, the model predicted only 1,142 pixels as belonging to the barren land class, while 1,644 pixels actually belonged to the barren land class.
Relying on such a data product may bias the model towards classes that are well-represented or well-classified according to the coarse resolution multitemporal datasets used to produce the land cover dataset, and against classes that are too small to be apparent in the coarse data (e.g., small barren clearings) or rely on phenological characteristics (e.g., crop fields that have been harvested prior to NAIP imagery acquisition).
Given the absence of literature on best practices for designing stratification protocols for sampling training data for patch-based classifiers, further analysis is needed to fully understand the implications of such approaches.

\section{Conclusion}

In this work, we have demonstrated that accurate land cover classification at VHSR can be achieved using as few as 1,000 256\(\times\)256-pixel patches of labeled data by utilizing a BYOL self-supervised pre-training objective to learn meaningful image representations from large amounts of unlabeled VHSR imagery prior to fully supervised fine-tuning.
We found that initializing a model with ImageNet weights and then continuing pre-training with BYOL was consistently the best-performing approach, suggesting that there is still value in utilizing models pre-trained on general-purpose imagery for remote sensing applications, even if there is a demonstrable domain gap.
Still, pre-training a model using BYOL from a random initialization was often competitive with ImageNet pre-training alone, demonstrating the power of in-domain self-supervision and the need for remote sensing-specific pre-training, even beyond the context of foundation models.
We used this approach to produce a \SIadj{1}{\meter} resolution land cover map for the entire state of Mississippi, USA, which captures fine-scale land cover features that are indistinguishable in existing land cover products such as the NLCD.
This achievement serves as a blueprint for both researchers applying geospatial deep learning models to data-scarce scenarios as well as practitioners seeking to produce high-quality land cover maps at VHSR over major geographic extents with limited human resources for collecting ground truth data.

\section*{Acknowledgments}

This work was supported by the Mississippi Space Grant Consortium (MSSGC) Graduate Fellowship Program through NASA funding, as well as by the Mississippi Agricultural and Forestry Experiment Station (MAFES) Special Research Initiative.
Dr. Martins and Lima were supported by the NASA Early Career Investigator Program in Earth Science (award \#80NSSC24K1040).
We also wish to acknowledge the University of Mississippi's Mississippi Center for Supercomputing Research and Mississippi State University's High Performance Computing Collaboratory for providing the computational resources necessary to complete this work.
An online dashboard to visualize and interact with the final land cover product can be found at \url{https://www.gcerlab.com/projects/msland}, while the final land cover product can be downloaded from \url{https://doi.org/10.5281/zenodo.16762344}.
The ground truth patches used to train the model can be found at \url{https://doi.org/10.5281/zenodo.15670823}.

\section*{CRediT authorship contribution statement}

\textbf{Dakota Hester}: Methodology, Software, Formal analysis, Data curation, Visualization, Writing - original draft.
\textbf{Vitor S. Martins}: Conceptualization, Methodology, Supervision, Writing - review \& editing, Funding acquisition, Project administration.
\textbf{Lucas B. Ferreira}: Methodology, Writing - review \& editing.
\textbf{Thainara M. A. Lima}: Writing - review \& editing.

\bibliographystyle{elsarticle-harv}
\bibliography{bibliography}

\clearpage

\section*{Supplementary material}

\setcounter{section}{0}
\setcounter{table}{0}
\setcounter{figure}{0}
\renewcommand{\thesubsection}{S\arabic{subsection}}
\renewcommand{\thetable}{S\arabic{table}}
\renewcommand{\thefigure}{S\arabic{figure}}

\subsection{Model performance breakdown}

\clearpage
\onecolumn

\begingroup
\scriptsize

\setlength\LTleft{0pt}
\setlength\LTright{0pt}

\begin{longtable}{@{\extracolsep{\fill}}lllcccccc}
\caption{\footnotesize Mean and standard deviation of accuracy metrics for all model configurations tested. User's accuracy, producer's accuracy, F1 score, and IoU are provided as macro-averaged metrics.}\label{tab:full_results}

\\

\toprule
Model & Pre-train Method & \(N_\text{train}\) & Overall Acc. (\%) & User's Acc. (\%) & Producer's Acc. (\%) & F1 Score (\%) & IoU (\%) & Cohen's \(\kappa\) \\
\midrule
\endfirsthead

\toprule
Model & Pre-train Method &  \(N_\text{train}\) & Overall Acc. (\%) & User's Acc. (\%) & Producer's Acc. (\%) & F1 Score (\%) & IoU (\%) & Cohen's k (\%) \\
\midrule
\endhead

\midrule
\endfoot

\bottomrule
\endlastfoot

Linear Probe
& ImageNet Only & 250 & $73.32 \pm 0.45$ & $53.01 \pm 3.75$ & $33.37 \pm 0.74$ & $35.01 \pm 1.30$ & $25.56 \pm 0.95$ & $0.4949 \pm 0.0134$ \\
& & 500 & $74.25 \pm 0.40$ & $54.78 \pm 2.15$ & $36.34 \pm 0.96$ & $38.08 \pm 1.38$ & $27.93 \pm 1.08$ & $0.5164 \pm 0.0092$ \\
& & 750 & $74.38 \pm 0.26$ & $55.22 \pm 0.97$ & $36.97 \pm 0.66$ & $38.68 \pm 1.04$ & $28.40 \pm 0.84$ & $0.5194 \pm 0.0059$ \\
\addlinespace
& MoCoV2 Only & 250 & $62.20 \pm 0.68$ & $21.36 \pm 6.22$ & $14.21 \pm 0.73$ & $12.57 \pm 1.17$ & $9.50 \pm 0.73$ & $0.1029 \pm 0.0311$ \\
& & 500 & $62.97 \pm 0.84$ & $27.31 \pm 5.52$ & $15.26 \pm 1.11$ & $14.25 \pm 1.72$ & $10.55 \pm 1.09$ & $0.1363 \pm 0.0373$ \\
& & 750 & $63.25 \pm 0.77$ & $32.79 \pm 7.91$ & $15.51 \pm 0.95$ & $14.66 \pm 1.53$ & $10.81 \pm 0.94$ & $0.1439 \pm 0.0271$ \\
\addlinespace
& ImageNet + MoCoV2 & 250 & $78.17 \pm 0.55$ & $64.91 \pm 2.96$ & $41.02 \pm 1.58$ & $43.81 \pm 2.15$ & $32.89 \pm 1.72$ & $0.5961 \pm 0.0126$ \\
& & 500 & $78.91 \pm 0.38$ & $62.18 \pm 0.46$ & $44.34 \pm 1.19$ & $47.01 \pm 1.48$ & $35.55 \pm 1.29$ & $0.6141 \pm 0.0080$ \\
& & 750 & $79.37 \pm 0.19$ & $59.47 \pm 0.42$ & $48.30 \pm 0.58$ & $50.05 \pm 0.56$ & $38.33 \pm 0.48$ & $0.6269 \pm 0.0030$ \\
\addlinespace
& BYOL Only & 250 & $80.11 \pm 0.42$ & $59.56 \pm 0.91$ & $50.46 \pm 1.03$ & $51.78 \pm 0.98$ & $40.21 \pm 1.14$ & $0.6451 \pm 0.0068$ \\
& & 500 & $80.65 \pm 0.33$ & $60.74 \pm 0.64$ & $51.87 \pm 0.48$ & $52.95 \pm 0.65$ & $41.47 \pm 0.66$ & $0.6547 \pm 0.0055$ \\
& & 750 & $80.91 \pm 0.24$ & $61.65 \pm 0.62$ & $52.43 \pm 0.39$ & $53.65 \pm 0.60$ & $42.10 \pm 0.46$ & $0.6600 \pm 0.0037$ \\
\addlinespace
& ImageNet + BYOL & 250 & $79.64 \pm 0.56$ & $60.45 \pm 0.89$ & $48.84 \pm 1.87$ & $50.71 \pm 2.28$ & $38.92 \pm 2.05$ & $0.6318 \pm 0.0114$ \\
& & 500 & $80.15 \pm 0.28$ & $59.93 \pm 0.76$ & $51.26 \pm 0.65$ & $52.59 \pm 0.79$ & $40.82 \pm 0.70$ & $0.6439 \pm 0.0047$ \\
& & 750 & $80.24 \pm 0.21$ & $59.91 \pm 0.61$ & $52.04 \pm 0.53$ & $53.23 \pm 0.59$ & $41.34 \pm 0.50$ & $0.6463 \pm 0.0041$ \\
\midrule
FCN
& ImageNet Only & 250 & $83.25 \pm 0.19$ & $72.00 \pm 0.33$ & $62.25 \pm 0.86$ & $64.44 \pm 0.68$ & $50.35 \pm 0.57$ & $0.7040 \pm 0.0037$ \\
& & 500 & $84.07 \pm 0.25$ & $73.17 \pm 1.02$ & $65.41 \pm 0.54$ & $67.07 \pm 0.52$ & $52.96 \pm 0.58$ & $0.7205 \pm 0.0042$ \\
& & 750 & $84.18 \pm 0.48$ & $72.90 \pm 1.11$ & $66.14 \pm 0.67$ & $67.38 \pm 0.94$ & $53.39 \pm 0.94$ & $0.7232 \pm 0.0080$ \\
\addlinespace
& MoCoV2 Only & 250 & $81.06 \pm 1.12$ & $66.23 \pm 2.93$ & $53.30 \pm 4.66$ & $56.54 \pm 3.15$ & $43.76 \pm 2.97$ & $0.6567 \pm 0.0273$ \\
& & 500 & $83.38 \pm 0.52$ & $69.49 \pm 1.53$ & $61.48 \pm 1.60$ & $63.02 \pm 1.50$ & $49.79 \pm 1.44$ & $0.7059 \pm 0.0096$ \\
& & 750 & $83.56 \pm 0.42$ & $71.15 \pm 0.90$ & $62.76 \pm 0.35$ & $64.15 \pm 0.32$ & $50.62 \pm 0.60$ & $0.7102 \pm 0.0065$ \\
\addlinespace
& ImageNet + MoCoV2 & 250 & $84.34 \pm 0.61$ & $72.05 \pm 1.07$ & $65.07 \pm 1.64$ & $66.52 \pm 1.47$ & $52.56 \pm 1.70$ & $0.7238 \pm 0.0105$ \\
& & 500 & $84.87 \pm 0.25$ & $73.08 \pm 0.48$ & $67.51 \pm 0.62$ & $68.44 \pm 0.52$ & $54.65 \pm 0.55$ & $0.7342 \pm 0.0039$ \\
& & 750 & $85.41 \pm 0.19$ & $74.11 \pm 0.22$ & $68.95 \pm 0.74$ & $69.83 \pm 0.29$ & $56.27 \pm 0.34$ & $0.7437 \pm 0.0031$ \\
\addlinespace
& BYOL Only & 250 & $83.10 \pm 0.25$ & $68.09 \pm 1.33$ & $60.94 \pm 0.92$ & $62.22 \pm 0.96$ & $48.62 \pm 0.90$ & $0.7037 \pm 0.0039$ \\
& & 500 & $84.03 \pm 0.41$ & $70.33 \pm 0.83$ & $63.64 \pm 0.52$ & $64.73 \pm 0.66$ & $51.07 \pm 0.67$ & $0.7199 \pm 0.0064$ \\
& & 750 & $84.20 \pm 0.23$ & $71.12 \pm 0.58$ & $64.86 \pm 0.76$ & $65.93 \pm 0.49$ & $52.07 \pm 0.37$ & $0.7239 \pm 0.0030$ \\
\addlinespace
& ImageNet + BYOL & 250 & $84.43 \pm 0.45$ & $72.05 \pm 0.68$ & $65.82 \pm 1.02$ & $66.93 \pm 1.14$ & $52.98 \pm 1.03$ & $0.7270 \pm 0.0079$ \\
& & 500 & $84.81 \pm 0.43$ & $73.12 \pm 0.41$ & $67.80 \pm 1.44$ & $68.59 \pm 1.03$ & $54.71 \pm 0.99$ & $0.7347 \pm 0.0072$ \\
& & 750 & $84.93 \pm 0.22$ & $74.03 \pm 0.77$ & $69.12 \pm 0.68$ & $69.67 \pm 0.52$ & $55.94 \pm 0.51$ & $0.7373 \pm 0.0030$ \\
\midrule
PAN
& ImageNet Only & 250 & $84.35 \pm 0.31$ & $74.20 \pm 0.92$ & $64.96 \pm 0.79$ & $67.55 \pm 0.73$ & $53.15 \pm 0.81$ & $0.7231 \pm 0.0061$ \\
& & 500 & $85.60 \pm 0.11$ & $76.13 \pm 0.79$ & $68.91 \pm 0.65$ & $70.74 \pm 0.68$ & $56.85 \pm 0.70$ & $0.7462 \pm 0.0016$ \\
& & 750 & $85.91 \pm 0.35$ & $77.23 \pm 0.34$ & $69.66 \pm 0.85$ & $71.44 \pm 0.69$ & $57.71 \pm 0.77$ & $0.7516 \pm 0.0063$ \\
\addlinespace
& MoCoV2 Only & 250 & $81.67 \pm 0.56$ & $69.85 \pm 2.31$ & $55.92 \pm 2.72$ & $59.56 \pm 1.78$ & $45.97 \pm 1.74$ & $0.6633 \pm 0.0123$ \\
& & 500 & $83.52 \pm 0.49$ & $73.04 \pm 1.46$ & $61.64 \pm 1.41$ & $64.20 \pm 0.90$ & $50.41 \pm 1.00$ & $0.7034 \pm 0.0093$ \\
& & 750 & $83.97 \pm 0.31$ & $74.12 \pm 1.56$ & $62.80 \pm 0.78$ & $64.96 \pm 0.82$ & $51.30 \pm 0.86$ & $0.7109 \pm 0.0062$ \\
\addlinespace
& ImageNet + MoCoV2 & 250 & $84.87 \pm 0.52$ & $73.12 \pm 1.22$ & $67.43 \pm 0.57$ & $68.60 \pm 0.84$ & $54.61 \pm 1.00$ & $0.7355 \pm 0.0083$ \\
& & 500 & $85.47 \pm 0.27$ & $75.43 \pm 0.57$ & $69.73 \pm 1.28$ & $70.92 \pm 0.78$ & $57.18 \pm 0.87$ & $0.7459 \pm 0.0056$ \\
& & 750 & $85.85 \pm 0.22$ & $76.51 \pm 0.99$ & $70.06 \pm 0.79$ & $71.54 \pm 0.41$ & $57.84 \pm 0.50$ & $0.7522 \pm 0.0039$ \\
\addlinespace
& BYOL Only & 250 & $83.64 \pm 0.33$ & $69.64 \pm 1.46$ & $63.35 \pm 0.73$ & $63.63 \pm 1.16$ & $49.76 \pm 0.96$ & $0.7140 \pm 0.0041$ \\
& & 500 & $84.13 \pm 0.55$ & $73.37 \pm 1.51$ & $64.86 \pm 1.09$ & $66.34 \pm 0.97$ & $52.37 \pm 0.97$ & $0.7224 \pm 0.0080$ \\
& & 750 & $84.72 \pm 0.66$ & $74.04 \pm 0.82$ & $65.53 \pm 1.27$ & $67.26 \pm 1.74$ & $53.45 \pm 1.67$ & $0.7327 \pm 0.0108$ \\
\addlinespace
& ImageNet + BYOL & 250 & $85.42 \pm 0.34$ & $75.44 \pm 1.07$ & $68.62 \pm 0.58$ & $69.93 \pm 0.85$ & $56.09 \pm 1.06$ & $0.7440 \pm 0.0051$ \\
& & 500 & $85.68 \pm 0.34$ & $76.43 \pm 1.33$ & $69.67 \pm 1.06$ & $71.06 \pm 0.89$ & $57.30 \pm 1.03$ & $0.7493 \pm 0.0057$ \\
& & 750 & $86.16 \pm 0.35$ & $77.23 \pm 0.19$ & $70.74 \pm 0.94$ & $71.88 \pm 0.53$ & $58.38 \pm 0.66$ & $0.7570 \pm 0.0062$ \\
\midrule
DeepLabV3+
& ImageNet Only & 250 & $84.51 \pm 0.18$ & $73.08 \pm 1.01$ & $67.50 \pm 1.00$ & $68.09 \pm 0.82$ & $53.93 \pm 0.83$ & $0.7281 \pm 0.0043$ \\
& & 500 & $85.20 \pm 0.23$ & $74.75 \pm 0.76$ & $70.08 \pm 0.71$ & $70.33 \pm 0.57$ & $56.40 \pm 0.65$ & $0.7416 \pm 0.0038$ \\
& & 750 & $85.54 \pm 0.50$ & $75.44 \pm 1.17$ & $71.76 \pm 1.26$ & $71.46 \pm 1.30$ & $57.74 \pm 1.52$ & $0.7473 \pm 0.0093$ \\
\addlinespace
& MoCoV2 Only & 250 & $81.11 \pm 1.57$ & $67.16 \pm 4.65$ & $53.08 \pm 6.49$ & $55.91 \pm 5.52$ & $42.73 \pm 5.04$ & $0.6548 \pm 0.0346$ \\
& & 500 & $83.22 \pm 0.24$ & $71.46 \pm 0.69$ & $62.65 \pm 1.14$ & $63.90 \pm 0.81$ & $49.81 \pm 0.86$ & $0.7028 \pm 0.0046$ \\
& & 750 & $83.55 \pm 0.46$ & $72.51 \pm 0.73$ & $63.01 \pm 2.36$ & $64.56 \pm 1.56$ & $50.32 \pm 1.74$ & $0.7076 \pm 0.0087$ \\
\addlinespace
& ImageNet + MoCoV2 & 250 & $84.91 \pm 0.71$ & $73.38 \pm 1.16$ & $69.54 \pm 1.67$ & $69.14 \pm 1.57$ & $55.20 \pm 1.72$ & $0.7360 \pm 0.0125$ \\
& & 500 & $85.68 \pm 0.35$ & $74.44 \pm 0.81$ & $71.94 \pm 1.23$ & $71.20 \pm 0.89$ & $57.56 \pm 1.02$ & $0.7505 \pm 0.0060$ \\
& & 750 & $86.01 \pm 0.12$ & $75.59 \pm 0.32$ & $72.50 \pm 0.31$ & $71.92 \pm 0.33$ & $58.35 \pm 0.32$ & $0.7556 \pm 0.0018$ \\
\addlinespace
& BYOL Only & 250 & $84.97 \pm 0.32$ & $72.98 \pm 1.20$ & $67.48 \pm 1.18$ & $68.01 \pm 1.32$ & $54.14 \pm 1.31$ & $0.7370 \pm 0.0057$ \\
& & 500 & $85.66 \pm 0.50$ & $74.35 \pm 0.94$ & $70.08 \pm 0.85$ & $70.04 \pm 1.15$ & $56.37 \pm 1.30$ & $0.7494 \pm 0.0087$ \\
& & 750 & $85.70 \pm 0.10$ & $74.88 \pm 0.73$ & $70.07 \pm 0.38$ & $70.39 \pm 0.36$ & $56.65 \pm 0.31$ & $0.7497 \pm 0.0021$ \\
\addlinespace
& ImageNet + BYOL & 250 & $85.55 \pm 0.32$ & $74.70 \pm 1.15$ & $70.21 \pm 0.57$ & $70.29 \pm 0.78$ & $56.46 \pm 0.98$ & $0.7469 \pm 0.0053$ \\
& & 500 & $86.21 \pm 0.30$ & $75.74 \pm 0.81$ & $72.66 \pm 0.83$ & $72.25 \pm 0.75$ & $58.74 \pm 0.88$ & $0.7593 \pm 0.0052$ \\
& & 750 & $86.39 \pm 0.18$ & $76.03 \pm 0.44$ & $73.32 \pm 0.72$ & $72.62 \pm 0.38$ & $59.20 \pm 0.40$ & $0.7627 \pm 0.0028$ \\
\midrule
UPerNet
& ImageNet Only & 250 & $84.66 \pm 0.34$ & $74.16 \pm 1.21$ & $68.20 \pm 0.90$ & $69.12 \pm 0.94$ & $54.73 \pm 1.15$ & $0.7311 \pm 0.0060$ \\
& & 500 & $85.32 \pm 0.33$ & $75.26 \pm 1.70$ & $69.87 \pm 1.94$ & $70.42 \pm 0.86$ & $56.44 \pm 0.92$ & $0.7433 \pm 0.0050$ \\
& & 750 & $85.58 \pm 0.39$ & $76.13 \pm 1.54$ & $71.21 \pm 0.60$ & $71.64 \pm 0.60$ & $57.82 \pm 0.73$ & $0.7485 \pm 0.0068$ \\
\addlinespace
& MoCoV2 Only & 250 & $81.54 \pm 0.97$ & $70.58 \pm 2.26$ & $57.48 \pm 3.05$ & $60.38 \pm 2.56$ & $46.36 \pm 2.65$ & $0.6620 \pm 0.0211$ \\
& & 500 & $84.51 \pm 0.54$ & $73.02 \pm 1.13$ & $64.52 \pm 2.04$ & $66.18 \pm 1.38$ & $52.43 \pm 1.55$ & $0.7245 \pm 0.0111$ \\
& & 750 & $84.18 \pm 0.63$ & $72.33 \pm 2.23$ & $65.37 \pm 0.99$ & $66.26 \pm 1.39$ & $52.22 \pm 1.77$ & $0.7204 \pm 0.0102$ \\
\addlinespace
& ImageNet + MoCoV2 & 250 & $82.23 \pm 3.62$ & $71.12 \pm 2.05$ & $69.35 \pm 1.07$ & $67.96 \pm 3.30$ & $54.24 \pm 2.85$ & $0.7004 \pm 0.0489$ \\
& & 500 & $84.00 \pm 1.02$ & $71.71 \pm 2.19$ & $71.63 \pm 1.32$ & $69.78 \pm 1.43$ & $55.93 \pm 1.70$ & $0.7260 \pm 0.0144$ \\
& & 750 & $83.92 \pm 1.41$ & $70.63 \pm 2.60$ & $72.32 \pm 0.23$ & $69.82 \pm 1.95$ & $55.69 \pm 2.47$ & $0.7263 \pm 0.0208$ \\
\addlinespace
& BYOL Only & 250 & $85.12 \pm 0.63$ & $72.52 \pm 1.17$ & $69.42 \pm 1.28$ & $69.08 \pm 1.18$ & $55.10 \pm 1.23$ & $0.7406 \pm 0.0098$ \\
& & 500 & $85.31 \pm 0.64$ & $73.52 \pm 2.57$ & $68.98 \pm 2.02$ & $69.54 \pm 2.16$ & $55.55 \pm 2.60$ & $0.7445 \pm 0.0112$ \\
& & 750 & $85.43 \pm 0.75$ & $74.65 \pm 1.12$ & $69.45 \pm 1.14$ & $69.87 \pm 1.06$ & $56.06 \pm 1.29$ & $0.7455 \pm 0.0126$ \\
\addlinespace
& ImageNet + BYOL & 250 & $85.90 \pm 0.55$ & $75.39 \pm 1.21$ & $70.41 \pm 1.75$ & $70.88 \pm 1.15$ & $57.23 \pm 1.30$ & $0.7529 \pm 0.0095$ \\
& & 500 & $86.19 \pm 0.45$ & $75.38 \pm 2.08$ & $72.94 \pm 0.43$ & $72.48 \pm 1.22$ & $58.85 \pm 1.55$ & $0.7600 \pm 0.0070$ \\
& & 750 & $86.59 \pm 0.29$ & $76.39 \pm 0.63$ & $73.06 \pm 0.21$ & $72.73 \pm 0.59$ & $59.31 \pm 0.82$ & $0.7655 \pm 0.0050$ \\
\midrule
Attention U-Net
& ImageNet Only & 250 & $84.90 \pm 0.23$ & $74.83 \pm 0.63$ & $69.71 \pm 0.66$ & $69.95 \pm 0.79$ & $56.05 \pm 0.82$ & $0.7347 \pm 0.0047$ \\
& & 500 & $85.91 \pm 0.47$ & $75.62 \pm 0.74$ & $72.96 \pm 1.00$ & $72.47 \pm 0.83$ & $58.80 \pm 0.92$ & $0.7542 \pm 0.0081$ \\
& & 750 & $86.16 \pm 0.38$ & $77.01 \pm 0.83$ & $73.21 \pm 0.77$ & $73.16 \pm 1.03$ & $59.59 \pm 1.32$ & $0.7589 \pm 0.0071$ \\
\addlinespace
& MoCoV2 Only & 250 & $84.20 \pm 0.43$ & $72.91 \pm 3.82$ & $61.92 \pm 1.93$ & $62.91 \pm 2.97$ & $49.72 \pm 2.03$ & $0.7182 \pm 0.0088$ \\
& & 500 & $85.61 \pm 0.49$ & $75.35 \pm 0.75$ & $67.82 \pm 1.45$ & $68.81 \pm 1.30$ & $54.98 \pm 1.43$ & $0.7462 \pm 0.0089$ \\
& & 750 & $85.56 \pm 0.74$ & $76.35 \pm 1.14$ & $67.31 \pm 4.04$ & $68.82 \pm 2.93$ & $55.05 \pm 2.71$ & $0.7439 \pm 0.0166$ \\
\addlinespace
& ImageNet + MoCoV2 & 250 & $85.35 \pm 0.71$ & $74.49 \pm 1.41$ & $71.07 \pm 1.61$ & $70.43 \pm 2.10$ & $56.56 \pm 2.43$ & $0.7439 \pm 0.0118$ \\
& & 500 & $86.23 \pm 0.54$ & $76.44 \pm 1.09$ & $73.11 \pm 0.81$ & $72.94 \pm 0.80$ & $59.49 \pm 0.97$ & $0.7597 \pm 0.0086$ \\
& & 750 & $86.37 \pm 0.25$ & $76.67 \pm 1.69$ & $73.61 \pm 0.35$ & $73.23 \pm 1.12$ & $59.83 \pm 1.24$ & $0.7629 \pm 0.0047$ \\
\addlinespace
& BYOL Only & 250 & $85.45 \pm 0.63$ & $73.33 \pm 1.47$ & $69.99 \pm 1.82$ & $69.78 \pm 0.93$ & $55.94 \pm 1.09$ & $0.7465 \pm 0.0098$ \\
& & 500 & $85.87 \pm 0.45$ & $74.86 \pm 0.68$ & $71.27 \pm 1.56$ & $71.02 \pm 1.38$ & $57.31 \pm 1.45$ & $0.7538 \pm 0.0081$ \\
& & 750 & $86.12 \pm 0.35$ & $74.95 \pm 0.45$ & $72.97 \pm 0.58$ & $71.70 \pm 0.45$ & $58.16 \pm 0.41$ & $0.7578 \pm 0.0061$ \\
\addlinespace
& ImageNet + BYOL & 250 & $85.93 \pm 0.58$ & $75.75 \pm 1.46$ & $71.90 \pm 0.68$ & $72.14 \pm 0.92$ & $58.33 \pm 1.15$ & $0.7559 \pm 0.0076$ \\
& & 500 & $86.32 \pm 0.29$ & $77.01 \pm 0.17$ & $73.28 \pm 0.94$ & $73.18 \pm 0.78$ & $59.65 \pm 0.91$ & $0.7618 \pm 0.0043$ \\
& & 750 & $86.65 \pm 0.41$ & $76.84 \pm 1.77$ & $75.15 \pm 0.35$ & $73.81 \pm 1.27$ & $60.55 \pm 1.52$ & $0.7677 \pm 0.0063$ \\
\midrule
U-Net
& ImageNet Only & 250 & $85.20 \pm 0.17$ & $75.55 \pm 0.48$ & $70.33 \pm 0.74$ & $70.62 \pm 0.48$ & $56.77 \pm 0.51$ & $0.7406 \pm 0.0036$ \\
& & 500 & $85.70 \pm 0.54$ & $76.48 \pm 1.36$ & $71.97 \pm 1.31$ & $72.03 \pm 1.34$ & $58.33 \pm 1.61$ & $0.7503 \pm 0.0088$ \\
& & 750 & $86.10 \pm 0.77$ & $77.43 \pm 1.99$ & $72.29 \pm 1.23$ & $72.57 \pm 1.01$ & $58.97 \pm 1.25$ & $0.7584 \pm 0.0115$ \\
\addlinespace
& MoCoV2 Only & 250 & $84.58 \pm 0.56$ & $74.50 \pm 2.03$ & $63.98 \pm 1.65$ & $65.27 \pm 1.68$ & $51.78 \pm 1.73$ & $0.7276 \pm 0.0093$ \\
& & 500 & $85.52 \pm 0.47$ & $75.36 \pm 1.03$ & $68.06 \pm 1.90$ & $68.86 \pm 1.66$ & $54.95 \pm 1.64$ & $0.7445 \pm 0.0089$ \\
& & 750 & $85.54 \pm 0.51$ & $74.79 \pm 1.17$ & $71.10 \pm 0.42$ & $70.49 \pm 0.76$ & $56.57 \pm 0.91$ & $0.7473 \pm 0.0079$ \\
\addlinespace
& ImageNet + MoCoV2 & 250 & $85.74 \pm 0.47$ & $75.90 \pm 1.10$ & $71.22 \pm 0.99$ & $71.41 \pm 1.33$ & $57.77 \pm 1.46$ & $0.7509 \pm 0.0079$ \\
& & 500 & $86.05 \pm 0.56$ & $76.24 \pm 2.31$ & $72.90 \pm 1.21$ & $72.63 \pm 1.69$ & $59.09 \pm 1.93$ & $0.7577 \pm 0.0094$ \\
& & 750 & $86.51 \pm 0.30$ & $77.51 \pm 0.56$ & $73.81 \pm 0.66$ & $73.79 \pm 0.70$ & $60.52 \pm 0.80$ & $0.7650 \pm 0.0050$ \\
\addlinespace
& BYOL Only & 250 & $85.34 \pm 0.29$ & $74.08 \pm 1.08$ & $69.91 \pm 0.99$ & $69.97 \pm 1.00$ & $56.13 \pm 1.02$ & $0.7444 \pm 0.0041$ \\
& & 500 & $86.20 \pm 0.45$ & $75.40 \pm 0.65$ & $71.66 \pm 1.65$ & $71.52 \pm 1.33$ & $57.90 \pm 1.49$ & $0.7593 \pm 0.0085$ \\
& & 750 & $86.25 \pm 0.27$ & $76.36 \pm 0.21$ & $72.19 \pm 0.47$ & $72.24 \pm 0.31$ & $58.71 \pm 0.28$ & $0.7602 \pm 0.0045$ \\
\addlinespace
& ImageNet + BYOL & 250 & $85.95 \pm 0.47$ & $76.91 \pm 0.88$ & $72.22 \pm 0.28$ & $72.45 \pm 0.51$ & $58.90 \pm 0.60$ & $0.7549 \pm 0.0075$ \\
& & 500 & $86.55 \pm 0.61$ & $77.46 \pm 0.85$ & $73.69 \pm 2.03$ & $73.70 \pm 1.71$ & $60.36 \pm 2.11$ & $0.7658 \pm 0.0105$ \\
& & 750 & $86.43 \pm 0.20$ & $78.18 \pm 0.27$ & $74.38 \pm 1.05$ & $74.04 \pm 0.86$ & $60.85 \pm 0.96$ & $0.7638 \pm 0.0043$ \\

\end{longtable}
\endgroup

\clearpage
\twocolumn

\subsection{Assessment of model misclassifications}~\label{sec:sup:error_assessment}

\begin{figure*}
    \centering
    \includegraphics[width=\textwidth]{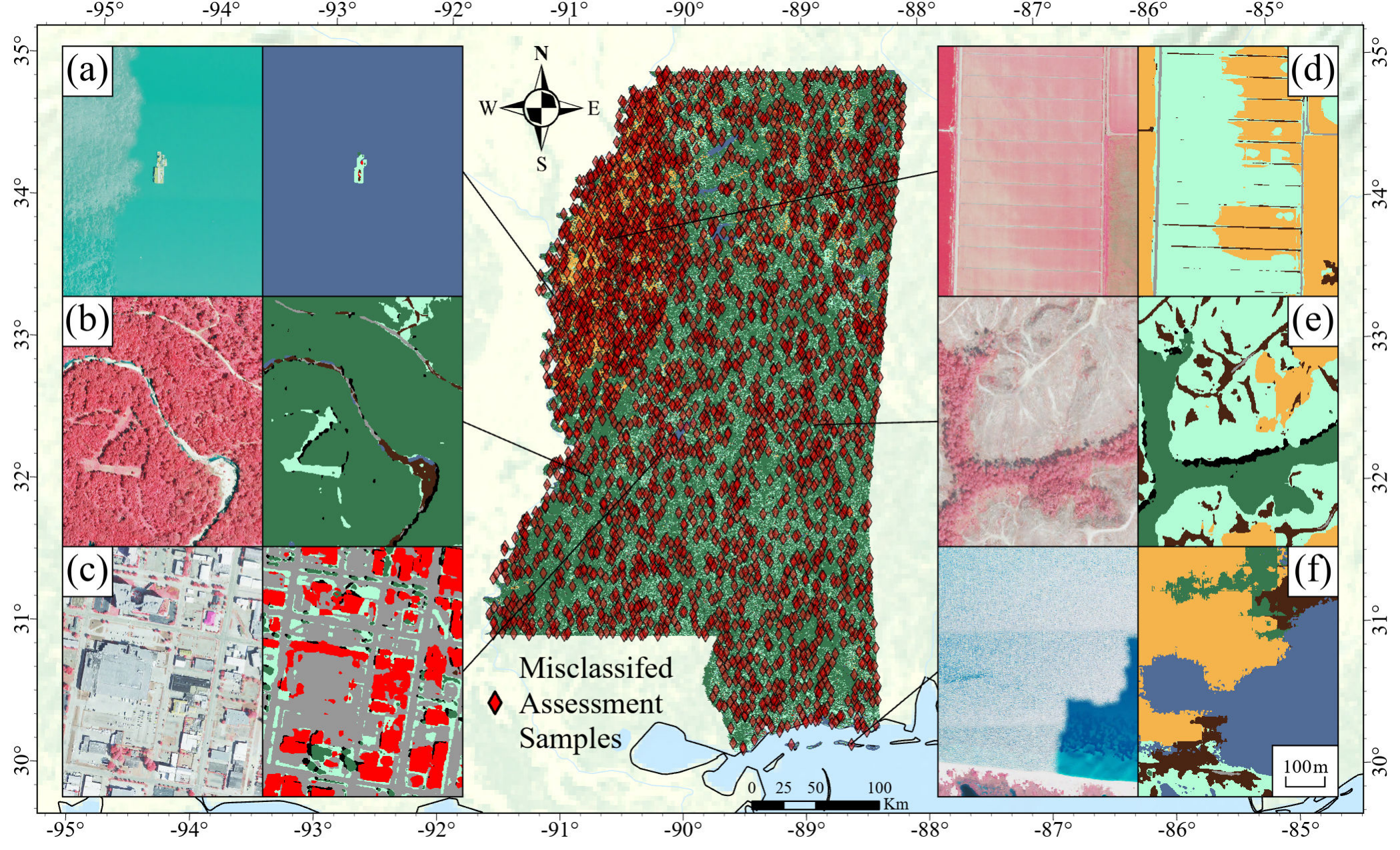}
    \caption{Spatial distribution of misclassified assessment samples in the final 2023 land cover product and examples of misclassifications.}\label{fig:error_analysis}
\end{figure*}

Examining the spatial distribution of errors in Figure~\ref{fig:error_analysis} reveals that errors are most heavily concentrated in the Mississippi Delta region, where agricultural activities are the predominant land use and a large amount of area is dedicated to cultivated crops.
Table~\ref{tab:class_wise_metrics} shows that our model has a high producer's accuracy for cultivated crops (95.54\%), but poor user's accuracy (65.91\%), indicating that while the model can accurately classify actual crop fields, it has a propensity to misclassify herbaceous land cover types as cultivated crops.
An example of such a misclassification is shown in samples (d) and (e) in Figure~\ref{fig:error_analysis}, where portions of herbaceous fields are misclassified as cropland.
The class definitions in Table~\ref{tab:class_legend} describes how both strong near-infrared reflectance (for crops near peak growth) and the presence of row patterns are potential indicators of cropland.
However, for crops where the spacing between rows is not visible in the imagery, or in cases where the imagery is acquired when the crop is not near peak growth, classification can be more complex.
To account for this, the model overclassified cultivated crop regions, leading to highly accurate classifications of regions that are actually cropland, but a tendency to commit errors of commission when classifying other vegetated land cover classes (e.g., herbaceous).

Unlike cultivated crops, the barren land class is prone to under-classification, with a moderately strong user's accuracy of 82.66\% but a weak producer's accuracy of 57.42\%.
Barren land regions are typically characterized by a lack of any texture and low near-infrared reflectance, indicating a total lack of vegetation.
However, for regions with mixed herbaceous/barren land cover types, the distinction between the two can be difficult to delineate precisely.
Sample (d) in Figure~\ref{fig:error_analysis} shows an example of this: while the herbaceous regions show some texture, it is difficult to visually distinguish where the herbaceous land cover ends and where the barren land begins.
Further, impervious surfaces are very similar to barren land cover types in terms of both spatial and spectral features.
Sample (b) in Figure~\ref{fig:error_analysis} demonstrates both errors of commission and omission with regards to the misclassification of barren land: the middle of the sample shows a small stream with barren banks and deposits, but narrow portions are misclassified as impervious surfaces, while the impervious road in the upper part of the image is partially misclassified as barren land.
In the absence of additional spectral information or ancillary road data, separating the two classes based on the color-infrared imagery alone is difficult.

From a numerical standpoint, classification of impervious structures (i.e., buildings) is moderately high with similar user's and producer's accuracy (76.19\% and 75.47\%, respectively).
Most misclassification of buildings appeared to arise from lack of nearby spatial features that aid in distinguishing impervious surfaces from impervious structures, such as shadows (indicating elevation) or hard edges/corners.
Sample (c) in Figure~\ref{fig:error_analysis} demonstrates the effect this has on classifications in dense urban landscapes, where large buildings have a tendency to be misclassified as impervious surfaces, while smaller buildings were classified more accurately.
This could be a limitation of the use of CNNs, which have limited receptive fields and struggle to incorporate long-range contextual features into classifications~\citep{luoUnderstandingEffectiveReceptive2016,liuUnderstandingEffectiveReceptive2018}.

Systematic errors arising from data quality issues are also present in the final product.
One such issue that is prevalent in rivers/waterways is the presence of shipping barges and other watercraft that cause misclassifications in open water areas. Such watercraft are mostly present in the Mississippi River due to its popularity as a shipping route.
Sample (a) in Figure~\ref{fig:error_analysis} shows an example of such an error: from a data product standpoint, watercraft are not permanent fixtures and such artifacts should be excluded from the model; however, without manually editing the dataset or masking large water bodies according to an external dataset, it is impossible for the model to handle watercraft appropriately.
This behavior is a byproduct of the mono-temporal nature of the input imagery: integrating multiple observations within a single year into the preprocessing pipeline or model inputs might address this problem, but doing so is impossible given the acquisition schedule of NAIP imagery.

Another systematic error is the appearance of sun glint in the imagery, leading to spurious misclassifications in the Gulf Coast region: sample (f) in Figure~\ref{fig:error_analysis} gives an example of such misclassifications in the Mississippi Sound region.
While quality control restricts the amount of allowable sun glint in NAIP imagery, it is impossible to account for and correct all instances of sun glint, leading to critical errors of omission when classifying water bodies.
Given that NAIP is rarely used for analysis of water and there is no widely used protocol for sun glint correction in NAIP imagery, these artifacts cannot be removed.
Including more samples of water bodies with high levels of sun glint in the training set is necessary to address this issue.

While the unclassified class provides little information from a data product perspective, its poor producer's accuracy (32.29\%) is potentially concerning.
The most likely cause for this is the fact that most unclassified pixels lie directly adjacent to or in the middle of forested regions due to the presence of shadows. While an annotator may strictly classify small groups of dark pixels as unclassified, the model tended to classify these pixels based on the adjacent land cover.
An example of this phenomenon can be observed in sample (b) in Figure~\ref{fig:error_analysis}, where small shadows in the forest are simply classified as belonging to the surrounding forest instead of the unclassified class, which is likely a correct classification given the context.
Table~\ref{tab:confusion_matrix} supports this conclusion, where it is evident that a large number of unclassified pixels were classified as tree canopy/woody vegetation.
While this is sub-optimal from a data quality standpoint, as the unclassified class has a moderately high user's accuracy (86.04\%) and the class provides no useful information on surface phenomena, it is unlikely that this has a large impact on the overall suitability of this product for a given task.

\subsection{2016 data product assessment}

\begin{table*}
  \centering
  \scriptsize
  \caption{Confusion matrix for the 2016 Mississippi land cover product.}\label{tab:confusion_matrix_2016}
  \begin{tabular*}{\textwidth}{@{\extracolsep{\fill}}lrrrrrrrrr}
    \toprule
    & \multicolumn{9}{l}{\textbf{Predicted Class}} \\
    \cmidrule(lr){2-10}
    \textbf{True Class} & \bfseries \specialcell[t]{Open\\Water} & \bfseries \specialcell[t]{Impervious\\Structures} & \bfseries \specialcell[t]{Impervious\\Surfaces} & \bfseries \specialcell[t]{Barren\\Land} & \bfseries \specialcell[t]{Tree Canopy/Woody\\Vegetation} & \bfseries \specialcell[t]{Low Vegetation/\\Herbaceous} & \bfseries \specialcell[t]{Cultivated\\Crops} & \bfseries Unclassified & \bfseries Total \\
    \midrule
    Open Water            & \textbf{565} & 1 & 2 & 10 & 30 & 35 & 5 & 24 & 672 \\
    Impervious Structures & 0 & \textbf{75} & 11 & 2 & 3 & 11 & 0 & 0 & 102 \\
    Impervious Surfaces   & 0 & 7 & \textbf{209} & 18 & 1 & 23 & 0 & 4 & 262 \\
    Barren Land           & 6 & 3 & 67 & \textbf{1,464} & 24 & 495 & 246 & 2 & 2,307 \\
    Tree Canopy           & 0 & 0 & 1 & 7 & \textbf{14,278} & 485 & 10 & 14 & 14,795 \\
    Low Vegetation        & 3 & 1 & 18 & 127 & 249 & \textbf{3,372} & 308 & 8 & 4,086 \\
    Cultivated Crops      & 0 & 0 & 1 & 43 & 0 & 101 & \textbf{793} & 0 & 938 \\
    Unclassified          & 6 & 7 & 11 & 6 & 1,239 & 130 & 0 & \textbf{439} & 1,838 \\
    \midrule
    Total                 & 580 & 94 & 320 & 1,677 & 15,824 & 4,652 & 1,362 & 491 & \textbf{25,000} \\
    \bottomrule
  \end{tabular*}
\end{table*}

\begin{table*}
  \centering
  \caption{Per-class classification metrics for the 2016 Mississippi land cover classification results.}\label{tab:class_wise_metrics_2016}
  \begin{tabular*}{\linewidth}{@{\extracolsep{\fill}}llllll}
    \toprule
    \bfseries Class Name & \bfseries User's Accuracy (\%) & \bfseries Producer's Accuracy (\%) & \bfseries F1 Score (\%) & \bfseries IoU (\%) & \bfseries Cohen's \(\kappa\) \\
    \midrule
    Open Water                & 97.41 & 84.08 & 90.26 & 82.24 & 0.9001 \\
    Impervious Structures     & 79.79 & 73.53 & 76.53 & 61.98 & 0.7644 \\
    Impervious Surfaces       & 65.31 & 79.77 & 71.82 & 56.03 & 0.7149 \\
    Barren Land               & 87.30 & 63.46 & 73.49 & 58.10 & 0.7126 \\
    Tree Canopy/Woody Vegetation & 90.23 & 96.51 & 93.26 & 87.38 & 0.8265 \\
    Low Vegetation/Herbaceous & 72.48 & 82.53 & 77.18 & 62.84 & 0.7237 \\
    Cultivated Crops          & 58.22 & 84.54 & 68.96 & 52.62 & 0.6751 \\
    Unclassified              & 89.41 & 23.88 & 37.70 & 23.23 & 0.3571 \\
    \midrule
    Macro Average             & 80.02 & 73.54 & 73.65 & 60.55 & 0.7093 \\
    Overall                   & Accuracy: 84.78 & & & 73.96 & 0.7751 \\
    \bottomrule
  \end{tabular*}
\end{table*}

\subsection{Pre-training dataset visualization}

\begin{figure}[H]
    \centering
    \includegraphics[width=\linewidth]{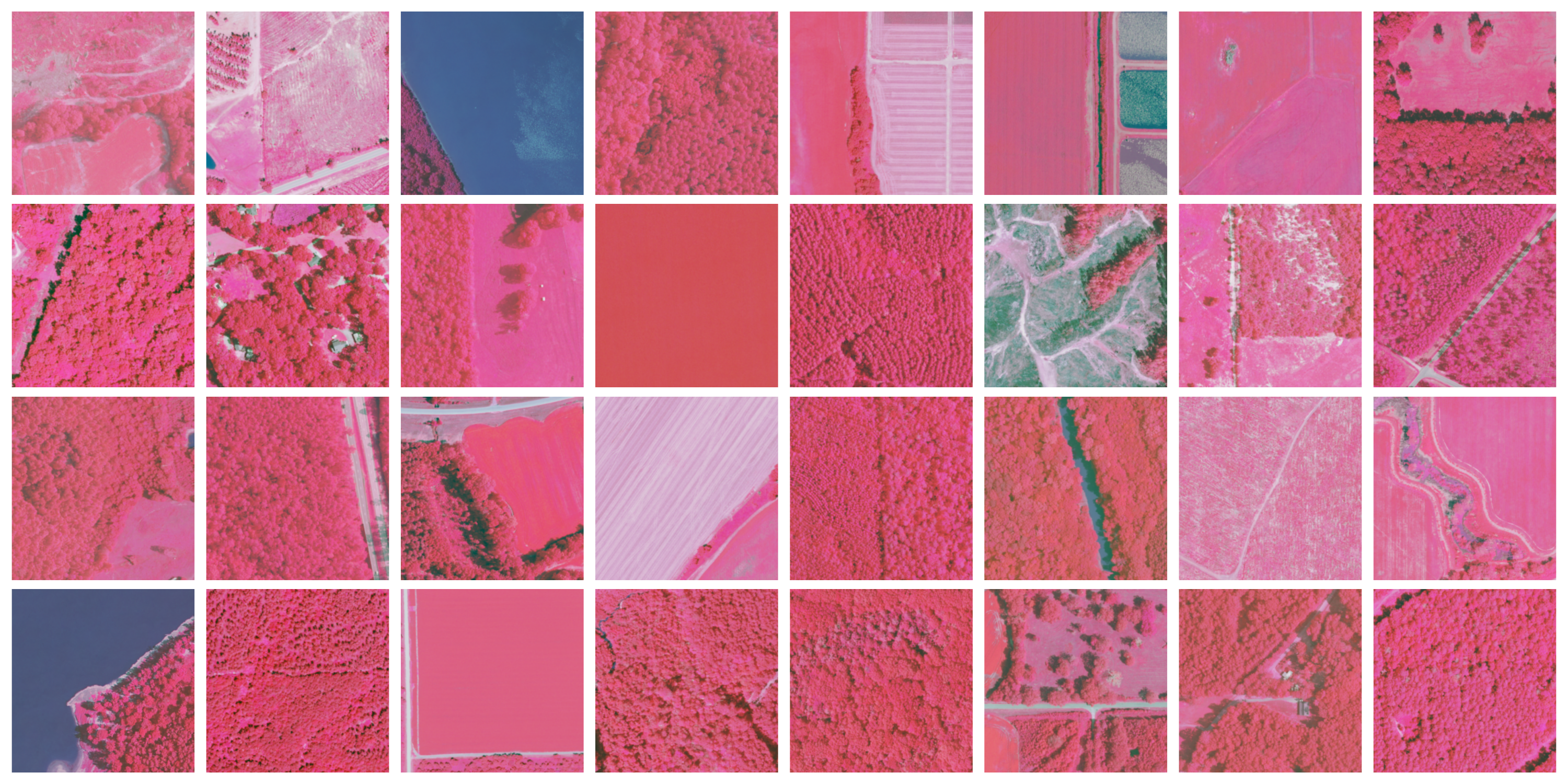}
    \caption{A random selection of samples in the pre-training dataset, demonstrating the homogeneity of landscapes throughout the pre-training dataset.}\label{fig:pretraining_samples}
\end{figure}

\end{document}